\documentclass[10pt,twocolumn,letterpaper]{article}

\usepackage[pagenumbers]{cvpr} 

\definecolor{cvprblue}{rgb}{0.21,0.49,0.74}
\usepackage[pagebackref,breaklinks,colorlinks,allcolors=cvprblue]{hyperref}
\usepackage{multicol}
\usepackage{multirow}
\usepackage{pifont}
\usepackage[table]{xcolor}
\usepackage[export]{adjustbox} 
\usepackage{makecell}

\newcommand{\ours}{Zoo3D}
\newcommand{\cmark}{\ding{51}}
\newcommand{\xmark}{\ding{55}}

\def\paperID{11196} 
\def\confName{CVPR}
\def\confYear{2026}

\title{\ours: \underline{Z}er\underline{o}-Sh\underline{o}t 3D Object Detection at Scene Level}

\author{Andrey Lemeshko\textsuperscript{1$\star$}, Bulat Gabdullin\textsuperscript{2$\star$}, Nikita Drozdov\textsuperscript{1}, Anton Konushin\textsuperscript{1}, \\ Danila Rukhovich\textsuperscript{3}, Maksim Kolodiazhnyi\textsuperscript{1\dag} \\ \textsuperscript{1}Lomonosov Moscow State University; \textsuperscript{2}Higher School of Economics; \\
\textsuperscript{3}M:3L Lab, Institute of Mechanics, Armenia}

\begin{document}
\maketitle

\begin{abstract}

\let\thefootnote\relax\footnotetext{\textsuperscript{$\star$}Equal contribution}\footnotetext{\textsuperscript{\dag}Corresponding author: kolodyazhniyma@my.msu.ru}
3D object detection is fundamental for spatial understanding. Real-world environments demand models capable of recognizing diverse, previously unseen objects, which remains a major limitation of closed-set methods. Existing open-vocabulary 3D detectors relax annotation requirements but still depend on training scenes, either as point clouds or images. We take this a step further by introducing \ours, the first training-free 3D object detection framework. Our method constructs 3D bounding boxes via graph clustering of 2D instance masks, then assigns semantic labels using a novel open-vocabulary module with best-view selection and view-consensus mask generation. \ours{} operates in two modes: the zero-shot \ours\textsubscript{0}, which requires no training at all, and the self-supervised \ours\textsubscript{1}, which refines 3D box prediction by training a class-agnostic detector on \ours\textsubscript{0}-generated pseudo labels. Furthermore, we extend \ours{} beyond point clouds to work directly with posed and even unposed images. Across ScanNet200 and ARKitScenes benchmarks, both \ours\textsubscript{0} and \ours\textsubscript{1} achieve state-of-the-art results in open-vocabulary 3D object detection. Remarkably, our zero-shot \ours\textsubscript{0} outperforms all existing self-supervised methods, hence demonstrating the power and adaptability of training-free, off-the-shelf approaches for real-world 3D understanding. Code is available at \url{https://github.com/col14m/zoo3d}.
\end{abstract}

\begin{figure}
\centering
\includegraphics[width=0.985\linewidth]{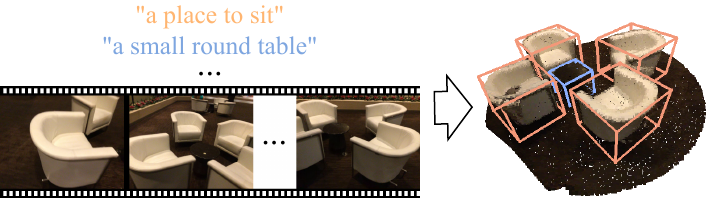}
\resizebox{\linewidth}{!}{
\begin{tabular}{lccccc}
\addlinespace[0.8em] 
\toprule
\multirow{2}{*}{Method} & \multicolumn{4}{c}{Seen during training} & ScanNet20 \\
& Depths & Images & Poses & Boxes & mAP\textsubscript{25} \\
\midrule
OV-Uni3DETR \cite{wang2024ov-uni3detr} & \cmark & \cmark & \cmark & \cmark & 25.3\\
OpenM3D \cite{hsu2025openm3d} & \cmark & \cmark & \cmark & & 19.8 \\
\rowcolor{blue!5} \ours\textsubscript{0} & & & & & 24.2 \\
\rowcolor{blue!5} \ours\textsubscript{1} & & \cmark & & & \textbf{27.9} \\
\bottomrule
\end{tabular}}
\caption{Open-vocabulary 3D object detection aims to localize 3D bounding boxes given a textual description. We demonstrate that this task can be solved in a zero-shot mode (\ours\textsubscript{0}). Our self-supervised image-based approach \ours\textsubscript{1} performs on par with point cloud-based methods that are trained with 3D bounding boxes supervision.}
\label{fig:teaser}
\end{figure}

\section{Introduction}

3D object detection aims to predict both category labels and oriented 3D bounding boxes for all objects within a scene. Adapting existing detection methods to real-world scenarios is often challenging and resource-intensive: collecting training data and fine-tuning models require substantial time and computational effort. Consequently, self-contained, off-the-shelf approaches with minimal supervision are highly desirable.

Fully supervised methods~\cite{qi2019votenet, rukhovich2022fcaf3d, rukhovich2023tr3d, kolodiazhnyi2025unidet3d, shen2023v-detr}, including recent LLM-based variants~\cite{zheng2025video-3d-llm, zhi2025lscenellm, zhu2024llava-3d, qi2025gpt4scene, mao2025spatiallm}, achieve impressive accuracy in closed-set 3D object detection but fail to generalize beyond training categories. Their performance remains limited by the diversity, size, and quality of annotated datasets. Self-supervised approaches~\cite{lu2023ov-3det, yang2024imov3d, wang2024ov-uni3detr, hsu2025openm3d} alleviate the need for manual annotations, yet their detection quality still lags behind fully supervised methods. Zero-shot detection represents the next frontier toward supervision-free 3D understanding. While zero-shot techniques have been proposed for related tasks such as 3D instance segmentation~\cite{yan2024maskclustering, tang2025onlineanyseg, zhao2025sam2object} and monocular 3D detection~\cite{yao2025labelany3d, yao2024ovmono3d}, multi-view 3D object detection has never been addressed in a truly zero-shot setting. Meanwhile, generalist vision–language models (VLMs)~\cite{bai2025qwen2.5-vl, guo2025seed1.5-vl} excel at single-view prediction and visual question answering from images or videos, but they still struggle to perform spatial reasoning at the scene level.

In 2D vision, zero-shot approaches~\cite{ren2024grounded-sam, liu2024grounding-dino} leverage foundation models such as SAM~\cite{ravi2024sam2} for segmentation and CLIP~\cite{radford2021learning} for visual–textual alignment. These same models have been repurposed for 3D understanding; for example, self-supervised 3D object detectors~\cite{hsu2025openm3d, yang2024imov3d} use CLIP and SAM to generate pseudo-labels during training. However, we ask a more ambitious question: \textit{can we develop a completely training-free 3D object detector?} In this work, we harness the potential of foundation models for training-free 3D understanding and present \ours\textsubscript{0}, the first-in-class zero-shot 3D object detection framework that outperforms existing self-supervised approaches. Moreover, we further improve prediction quality with a self-supervised \ours\textsubscript{1}.

Beyond training supervision, inference requirements also pose challenges. Point clouds—commonly required by 3D understanding methods—are not always available in practice. We overcome this limitation by employing another foundation model, DUSt3R~\cite{wang2024dust3r}, to bridge the gap between 2D and 3D representations.  This way, we enable 3D object detection directly from images, which is a step toward democratizing spatial intelligence.

In summary, we investigate how far the requirements for 3D object detection can be relaxed. Starting from point clouds and ground-truth annotations, we progressively reduce supervision and input modality requirements, ultimately introducing a training-free approach that operates directly on unposed images. Even in this most constrained setting, our method performs on par with existing approaches that rely on extensive training and point cloud inputs at inference time. Our contributions are as follows:
\begin{itemize}
\item We define a novel task: zero-shot indoor 3D object detection at the scene level.
\item We propose novel methods for open-vocabulary 3D object detection in zero-shot and self-supervised settings.
\item Our methods achieve state-of-the-art results on four ScanNet-based benchmarks across three input modalities: point clouds, posed images, and unposed images.
\end{itemize}

\section{Related Work}

\paragraph{3D Object Detection from Point Clouds.}
Existing methods for 3D object detection from point clouds can be categorized by the level of supervision involved. 

Supervised methods, such as FCAF3D~\cite{rukhovich2022fcaf3d}, TR3D~\cite{rukhovich2023tr3d}, UniDet3D~\cite{kolodiazhnyi2025unidet3d}, and V-DETR~\cite{shen2023v-detr}, rely on fully annotated 3D bounding boxes. More recently, LLM-based approaches like Video-3D LLM~\cite{zheng2025video-3d-llm}, LSceneLLM~\cite{zhi2025lscenellm}, and Chat-Scene~\cite{huang2024chat-scene} have leveraged multimodal reasoning to enhance semantic understanding, but still are fundamentally constrained by the limited size and diversity of annotated 3D datasets, and cannot generalize beyond seen classes.

Semi-supervised approaches, e.g., OV-Uni3DETR~\cite{wang2024ov-uni3detr}, CoDa~\cite{cao2023coda}, OV-3DETIC~\cite{lu2022ov-3detic}, Object2Scene~\cite{zhu2023object2scene}, INHA~\cite{jiao2024inha}, and FM-OV3D~\cite{zhang2024fm-ov3d}, are trained on a subset of labeled 3D bounding boxes and evaluated on both seen and unseen categories in an open-vocabulary mode. These models bridge the gap between closed-set and open-world detection but still rely on training scenes with ground truth boxes.

Self-supervised methods, including OV-3DET~\cite{lu2023ov-3det}, ImOV3D~\cite{yang2024imov3d}, and GLIS~\cite{peng2024glis}, eliminate the need for explicit 3D labels by generating pseudo-annotations or distilling 2D supervision. Despite reduced supervision, they still require exposure to 3D scenes and rely on suboptimal pseudo-box generation or text-visual alignment strategies. Our method improves this stage via a simplified and training-free pipeline using only CLIP~\cite{radford2021learning} and SAM~\cite{ravi2024sam2}.

Zero-shot approaches do not require any data from training scenes. Outdoor-focused methods such as SAM3D~\cite{zhang2023sam3d-outdoor} operate on BEV projections but do not generalize to indoor scenes. MaskClustering~\cite{yan2024maskclustering}, OnlineAnySeg~\cite{tang2025onlineanyseg}, and SAM2Object~\cite{zhao2025sam2object} address open-vocabulary 3D instance segmentation, yet not 3D object detection. To our knowledge, we are the first to formulate and tackle indoor 3D object detection in the zero-shot regime.

\paragraph{3D Object Detection from Posed Multi-view Images.}
Supervised methods such as ImVoxelNet~\cite{rukhovich2022imvoxelnet}, ImGeoNet~\cite{tu2023imgeonet}, and NeRF-Det++~\cite{huang2025nerf-det++} derive closed-vocabulary 3D scene representation with 3D objects from collections of images with corresponding poses. Recent LLM-powered models like LLaVA-3D~\cite{zhu2024llava-3d} and GPT4Scene~\cite{qi2025gpt4scene} introduce large-scale vision-language understanding but still require labeled training data.

In the open-vocabulary regime, OV-Uni3DETR~\cite{wang2024ov-uni3detr} unifies training across both images and point clouds, while OpenM3D~\cite{hsu2025openm3d} combines CLIP features with voxelized 3D representations. The latter is most related to our approach but depends on depth supervision and CLIP finetuning, whereas we remain entirely training-free and use frozen CLIP features only at inference, yielding stronger performance. No prior methods address zero-shot 3D object detection directly from posed images; \ours\textsubscript{0} is the first-in-class method tackling this task.

\paragraph{3D Object Detection from Unposed Multi-view Images.}
The most unconstrained setting, unposed multi-view images, has only recently been explored by LLM-based approaches such as VLM-3R~\cite{fan2025vlm-3r} and SpatialLM~\cite{mao2025spatiallm}, which require full supervision. No prior work has attempted open-vocabulary, self-supervised, or zero-shot 3D object detection in this modality. Our method fills this gap by leveraging DUSt3R~\cite{wang2024dust3r} for pose-free reconstruction, enabling both self-supervised and zero-shot detection.

\begin{figure*}
\centering
\includegraphics[width=0.985\linewidth]{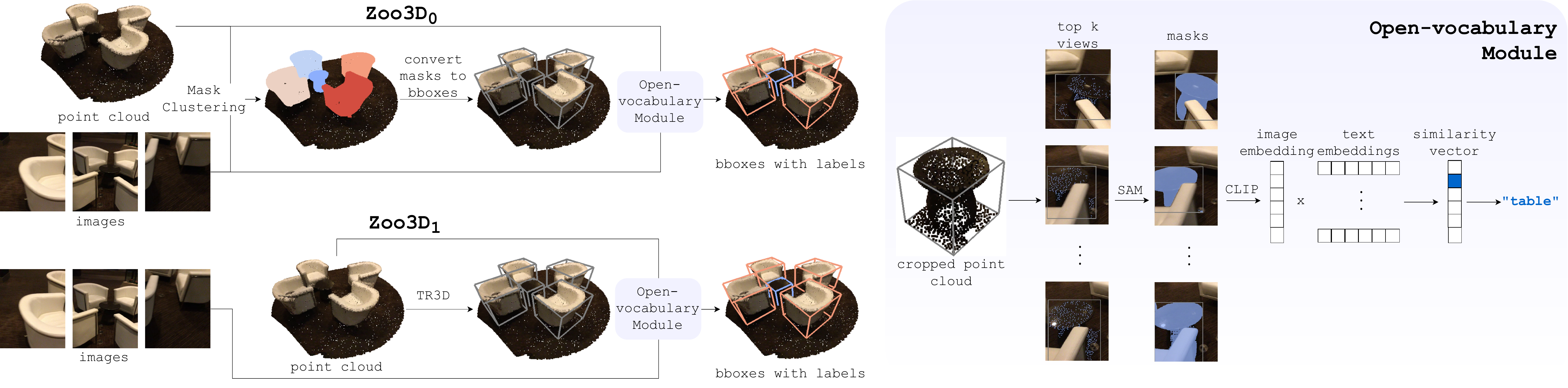}
\caption{Inference pipeline of \ours~ given point cloud inputs. \ours\textsubscript{0} leverages MaskClustering to predict class-agnostic 3D bounding boxes from a point cloud and images (top-left), while \ours\textsubscript{1} infers 3D bounding boxes from point clouds with TR3D (bottom-left). Both \ours\textsubscript{0} and \ours\textsubscript{1} assign semantic labels to 3D bounding boxes with the same Open-vocabulary Module (right). Given images, a full point cloud, and a 3D bounding box of an object, it crops the point cloud using the 3D bounding box, selects top k views based on the visibility, and projects visible points of the object onto these views. Object masks are obtained with SAM, CLIP embeddings are aggregated across views, and the text label with the most similar embedding is assigned.}
\label{fig:scheme}
\end{figure*}

\section{Open-vocabulary 3D Object Detection from Point Clouds}

Our work targets open-vocabulary 3D object detection, where the model must localize and recognize arbitrary objects. The open-vocabulary setting implies the absence of ground-truth 3D bounding box annotations. Recent annotation-free approaches exploit alternative supervision such as point clouds, RGB images, and camera trajectories, often leveraging powerful foundation models (e.g., CLIP, SAM) to align visual and text modalities. While these methods reduce dependence on manual labels, they still require access to training data.

To the best of our knowledge, no existing method can perform open-vocabulary 3D object detection entirely in a training-free manner, without using any form of training data. We are the first to address this challenge in a zero-shot setting. Below, we formalize the problem in Sec.~\ref{ssec:problem-formulation}, and present our zero-shot framework, \ours\textsubscript{0}, in Sec.~\ref{ssec:zero-shot}. Furthermore, we extend these ideas to a self-supervised variant, \ours\textsubscript{1}, in Sec.~\ref{ssec:self-supervised}, which leverages unlabeled data to enhance detection performance.

\subsection{Problem Formulation} \label{ssec:problem-formulation}

3D object detection implies estimating 3D bounding boxes of individual objects, given inputs in a form of a point cloud, images, depths, camera poses, camera intrinsics, or any combination of them. Particularly, \ours{} model awaits a point cloud $\mathcal{P}=\{p_i\}_{i=1}^N \subset \mathbb{R}^3$, where each point $p_i=(x_i,y_i,z_i)$ is described with its coordinates in the 3D space, color images $\mathcal{I}=\{I_1, I_2, \ldots, I_T \}$, their corresponding depths $\mathcal{D}=\{D_1, D_2, \ldots, D_T \}$, camera extrinsics  $\mathcal{R}=\{R_1, R_2, \ldots, R_T \}$ and intrinsic $K$. 3D objects are parameterized as $\mathcal{O}=\{(b_g,  l_g)\}_{g=1}^{G}$, where $l_g\in\{1,\dots,L\}$ denotes object open-vocabulary labels and $b_g$ stands for the spatial parameters of a 3D bounding box. $b_g=(c_g,s_g)$, where $c_g\in\mathbb{R}^3$ is the center of a 3D bounding box and $s_g\in\mathbb{R}^3_{+}$ are sizes along the $x,y,z$-axes.

\subsection{Zero-shot 3D Object Detection: \ours\textsubscript{0}} \label{ssec:zero-shot}

In \ours{}, we decompose open-vocabulary 3D object detection into class-agnostic 3D object detection and open-vocabulary label assignment. First, we predict class-agnostic 3D bounding boxes $\{b_g\}_{g=1}^{G}$. Then, each box gets assigned with a semantic label $l_g$ in our novel open-vocabulary module.

\paragraph{Class-agnostic 3D object detection}

is built upon zero-shot 3D instance segmentation method, namely, state-of-the-art MaskClustering~\cite{yan2024maskclustering}. The original approach operates on color images $\mathcal{I}$, depths $\mathcal{D}$, camera extrinsics $\mathcal{R}$ and a reconstructed point cloud $\mathcal{P}$ and returns points with instance labels. We extend this approach so that it outputs 3D bounding boxes, thereby serving as a class-agnostic 3D object detection method. 

The pipeline is organized as follows. First, a class-agnostic mask predictor produces 2D masks $\{m_{t,i}\}$ for each frame. These masks are used as nodes in a mask graph, with edges connecting masks belonging to the same instance. This graph is built iteratively by adding edges between highly conformed masks. For mask $m_{t',i}$ in the frame $t'$ and mask $m_{t'',j}$ in the frame $t''$, we calculate \textit{view consensus rate} $\text{cr}(m_{t',i}, m_{t'',j})$ as a measure of conformity. To this end, we find all \textit{observer} frames $F_o$ where both masks $m_{t',i}$ and $m_{t'',j}$ are visible. Among those views, we find \textit{supporter} frames $F_s \in F_o$: a view $t$ is a supporter if it contains a mask $m_{t,k}$ whose point cloud $P_{t,k}$ includes the visible portions from point clouds $P_{t',i}^t, P_{t'',j}^t$ of both $m_{t',i}$ and $m_{t'',j}$. The consensus rate is the supporter-observer ratio:
\[
\text{cr}(m_{t',i}, m_{t'',j}) = \frac{|\{t \in V \mid \exists k, P_{t',i}^t, P_{t'',j}^t \sqsubset P_{t,k}\}|}{|F_o|}.
\]

An edge is created if $\text{cr}(m_{t',i}, m_{t'',j}) \geq \tau_{rate}=0.9$. For each mask $m_{t',i}$, denoting the set of all masks that contain it across views as $M(m_{t',i})$, and its set of visible frames $F(m_{t',i})$, the consensus rate can be rewritten as:
\[
\text{cr}(m_{t',i}, m_{t'',j}) = \frac{|M(m_{t',i}) \cap M(m_{t'',j})|}{|F(m_{t',i}) \cap F(m_{t'',j})|}.
\]

\begin{figure*}
\begin{minipage}[t]{0.37\linewidth}
\vspace{0pt} \centering
\begin{tabular}{lcc}
\toprule
\multirow{2}{*}{Method} & \multicolumn{2}{c}{Seen during training}  \\
& Scenes & 3D bounding boxes \\
\midrule
\ours\textsubscript{0} & \xmark & \xmark \\
\ours\textsubscript{1} & \cmark & generated w/ \ours\textsubscript{0} \\
\ours\textsubscript{2} & \cmark & generated w/ \ours\textsubscript{1} \\
\bottomrule
\end{tabular}
\captionof{table}{Data used for training \ours{} models.}
\label{tab:zoo3d012}
\end{minipage}
\hfill
\begin{minipage}[t]{0.6\linewidth}
\vspace{0pt} \centering
\centering
\includegraphics[width=\linewidth]{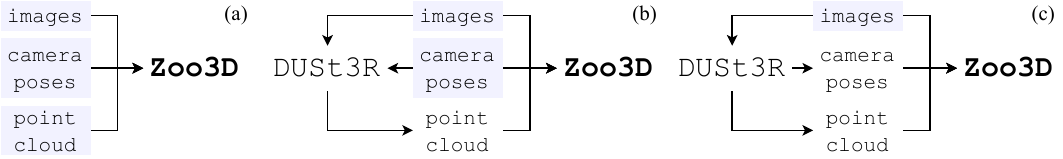}
\captionof{figure}{Three operation modes of \ours{}: with images with corresponding camera poses and point clouds as inputs (a), posed images (b) and unposed images (c). Ground truth input modalities are marked blue. In two latter scenarios, missing modalities are derived using DUSt3R.}
\label{fig:dust3r}
\end{minipage}
\end{figure*}

Masks are merged iteratively. In each iteration $k$, we remove edges with less than $n_k$ observers, and merge connected components into new nodes. The final 3D instance masks are turned into 3D bounding boxes. Since we consider only axis-aligned boxes without rotation, the task boils down to getting minimum and maximum coordinates of 3D points belonging to each mask. 

\paragraph{Open-vocabulary module}

assigns class-agnostic 3D bounding boxes with open-vocabulary semantic labels to each bounding box. First, we crop a point cloud $\mathcal{P}$ with a bounding box $b_g$:
\[
\mathcal{P}_g = \{ p_i \in \mathcal{P} \mid c_g - \frac{s_g}{2} \le p_i \le c_g + \frac{s_g}{2} \}.
\]
Using the camera intrinsic and extrinsics, we project the 3D points from $\mathcal{P}_g$ onto the 2D images. The projection of a 3D point $p_i$ onto frame $t$ is given by:
\[
u_{t,i} = \pi(K R_t [x_i, y_i, z_i, 1]^T).
\]
where $\pi$ is the perspective projection operation $\pi([x, y, w]^T) = [\frac{x}{w}, \frac{y}{w}]$.

Next, we backproject points and filter out significantly displaced points to avoid occlusions: 
\[
a_{t, i} = (K^{-1} [u_{t,i}, 1]^T) \cdot D_t(u_{t,i}),
\]
\[
\mathcal{U}_g^t = \left\{ u_{t,i} \ \middle| \ \| p_i - (R_t^{-1} [a_{t, i}, 1]^T)_{x,y,z}  \| < \tau_\text{occ} \right\}.
\]
where $\mathcal{U}_g^t$ -- filtered points of bounding box $b_g$ projected on image $I_t$,  $\tau_\text{occ}$ is the occlusion threshold.

We then select the top five images with the largest number of projected points. Since the predicted bounding boxes have some inherent error, the projection of the mask $\mathcal{U}_g^t$ may not align perfectly with the object in the image. To solve this problem, we frame the projected points with a 2D bounding box using a min/max operation:
\[
\text{bb}_{2d} = [\min(\mathcal{U}_{g,x}^t), \min(\mathcal{U}_{g,y}^t), \max(\mathcal{U}_{g,x}^t), \max(\mathcal{U}_{g,y}^t)].
\]
This 2D bounding box serves as an initial prompt for SAM, which returns a refined segmentation mask. This mask is then processed at three different scales, as in MaskClustering, and fed into CLIP to obtain a feature vector. Finally, the average feature vector across five images and three scales is used to compute the cosine similarity with the feature vectors of the class names, thus yielding the labels and confidences for an arbitrary set of classes.

\subsection{Self-supervised 3D Object Detection: \ours\textsubscript{1}} \label{ssec:self-supervised}

Our training-free solution already sets a state-of-the-art even compared with self-supervised methods. Next, we explore how the accuracy can be pushed even higher utilizing our zero-shot model as a basis. Specifically, we run \ours\textsubscript{0} to generate class-agnostic 3D bounding boxes for training scenes, and train a class-agnostic 3D object detection model on the obtained labeled dataset in a self-supervised mode. 

\paragraph{Class-agnostic 3D object detection model}

We build upon 3D object detection TR3D~\cite{rukhovich2023tr3d} and introduce modifications to adapt it to open vocabulary scenario. input point clouds are voxelized into 2 cm voxels and processed with a 3D sparse ResNet~\cite{rukhovich2022fcaf3d} that transforms the voxel space into 8 cm, 16 cm, 32 cm, and 64 cm-sized spatial grids. Neck aggregates 3D voxel features from four residual levels of the backbone. at levels of 64 cm and 32 cm, generative convolutional layers are added in order to prevent information loss and preserve the visibility field. 

The detection head consists of two stacked linear layers. TR3D  splits objects of interest into "large" and "small" based on their category, and predicts objects of each size with a dedicated head: "large" objects are predicted at 32-cm level, while "small" objects are inferred at 16-cm level. since we operate in open vocabulary mode, we cannot predefine any category-based routing. Accordingly, we use only the 16 cm-level. 

Besides, we do not need to predict the object category but just estimate probability of object's presence, we omit classification-related part of the original model. In our class-agnostic model, the head returns a set of 3D locations $\hat{\mathcal V}=\{\hat v_j\}_{j=1}^{J}$. for each location $\hat v_j\in\hat{\mathcal V}$, it returns an objectness logit $\tilde z_j\in\mathbb{R}^{1}$, an offset of the center of an object $\Delta c_j\in\mathbb{R}^{3}$, and log-sizes of its 3D bounding box $\tilde s_j\in\mathbb{R}^{3}$. The canonical representation of the predicted 3D bounding box is derived as $c_j = \hat v_j + \Delta c_j$, $s_j = \exp(\tilde s_j) \in \mathbb{R}^3_{+}$,
while the resulting class probabilities are calculated as $p_{j}=\sigma(\tilde z_{j})$.

\paragraph{Training procedure}

The assigner is used to couple the 3D locations ${\hat{v}_j}$ with the ground truth objects $\mathcal{O}_\text{gt}$. Each ground truth object is assigned to the six nearest locations to its center. The loss function is bi-component, with components corresponding to head outputs. Namely, objectness prediction in the 3D object detection head is guided with a focal loss $\mathcal{L}_{\text{focal}}$ and regression of 3D bounding box parameters is being trained with DIoU loss $\mathcal{L}_{\text{DIoU}}$:
\[
\mathcal{L} = \mathcal{L}_{\text{focal}} + \mathcal{L}_{\text{DIoU}}.
\]

\begin{table*}
\centering
\begin{tabular}{llcccccccc}
\toprule
& \multirow{2}{*}{Method} & \multirow{2}{*}{Venue} & Zero- & \multicolumn{2}{c}{ScanNet20}  & \multicolumn{2}{c}{ScanNet60} & \multicolumn{2}{c}{ScanNet200} \\
& & & shot & mAP\textsubscript{25} & mAP\textsubscript{50} & mAP\textsubscript{25} & mAP\textsubscript{50} & mAP\textsubscript{25} & mAP\textsubscript{50} \\
\midrule
\multicolumn{7}{l}{\textit{Point cloud + posed images}} \\
& Det-PointCLIPv2\textsuperscript{\dag}~\cite{zhu2023pointclip} & ICCV'23 & \xmark & - & - & 0.2 & - & - & - \\
& 3D-CLIP\textsuperscript{\dag}~\cite{radford2021learning} & ICML'21 & \xmark & - & - & 3.9 & - & - & - \\
& OV-3DET~\cite{lu2023ov-3det} & CVPR'23 & \xmark & 18.0 & - & - & - & - & - \\
& CoDa\textsuperscript{\dag}~\cite{cao2023coda} & NIPS'23 & \xmark & 19.3 & - & 9.0 & - & - & - \\
& INHA\textsuperscript{\dag}~\cite{jiao2024inha} & ECCV'24 & \xmark & - & - & 10.7 & - & - & - \\
& GLIS~\cite{peng2024glis} & ECCV'24 & \xmark & 20.8 & - & - & - & - & - \\
& ImOV3D~\cite{yang2024imov3d} & NIPS'24 & \xmark & 21.5 & - & - & - & - & - \\
& OV-Uni3DETR\textsuperscript{\dag}~\cite{wang2024ov-uni3detr} & ECCV'24 & \xmark & 25.3 & - & 19.4 & - & - & - \\
\rowcolor{blue!5} & \ours\textsubscript{0} & - & \cmark & 34.7 & 23.9 & 27.1 & 18.7 & 21.1 & 14.1 \\
\rowcolor{blue!5} & \ours\textsubscript{1}  & - & \xmark & \textbf{37.2} & \textbf{26.3} & \textbf{32.0} & \textbf{20.8} & \textbf{23.5} & \textbf{15.2} \\
\midrule
\multicolumn{7}{l}{\textit{Posed images}} \\
& OV-Uni3DETR\textsuperscript{\dag}~\cite{wang2024ov-uni3detr} & ECCV'24 & \xmark  &  -  &  - & 11.2 & - & - & - \\
& SAM3D~\cite{yang2023sam3d} $\rightarrow$ OpenM3D~\cite{hsu2025openm3d} & ICCV'25 & \xmark  & 16.7  &  5.2 & - & - & 3.9 & - \\
& OV-3DET~\cite{lu2023ov-3det} $\rightarrow$ OpenM3D~\cite{hsu2025openm3d} & ICCV'25 & \xmark  & 17.7 & 2.9  & - & - & 3.1 & - \\
& OpenM3D~\cite{hsu2025openm3d} & ICCV'25 & \xmark  & 19.8  & 7.3 & - & - & 4.2 & - \\
\rowcolor{blue!5} & DUSt3R~\cite{wang2024dust3r} $\rightarrow$ \ours\textsubscript{0} & - & \cmark & 30.5 & \textbf{17.3} & 22.0 & 10.4 & 14.3 & 6.2 \\
\rowcolor{blue!5} & DUSt3R~\cite{wang2024dust3r} $\rightarrow$ \ours\textsubscript{1} & - & \xmark & \textbf{32.8} & 15.5 & \textbf{23.9} & \textbf{10.8} & \textbf{16.5} & \textbf{6.3} \\
\midrule
\multicolumn{7}{l}{\textit{Unposed images}} \\
\rowcolor{blue!5} & DUSt3R \cite{wang2024dust3r} $\rightarrow$ \ours\textsubscript{0} & - & \cmark & 24.2 & 8.8 & 13.3 & 4.1 & 8.3 & 2.9 \\
\rowcolor{blue!5} & DUSt3R \cite{wang2024dust3r} $\rightarrow$ \ours\textsubscript{1} & - & \xmark & \textbf{27.9} & \textbf{10.4} & \textbf{15.3} & \textbf{5.6} & \textbf{10.7} & \textbf{3.8} \\
\bottomrule
\end{tabular}
\caption{Results of open-vocabulary 3D object detection on the ScanNet dataset across three benchmarks (with 20, 60, and 200 classes), and three input modalities (points cloud, posed and unposed multi-view images). Methods with \textsuperscript{\dag} utilize 3D bounding box annotations during training. Our \ours{} significantly outperforms prior work even in the zero-shot setting. Self-supervised \ours\textsubscript{1} using only unposed images performs on par with point cloud- based methods.}
\label{tab:results-scannet}
\end{table*}

\vspace*{-2em}
\paragraph{Iterative training}

\ours\textsubscript{1} is trained with class-agnostic 3D bounding boxes generated by \ours\textsubscript{0}. Experiments show that the first iteration of this procedure improves the quality. Guided by this observation, we perform the second iteration of the procedure. Specifically, we re-generate annotations now using \ours\textsubscript{1}, and reproduce the training procedure with the same model and the same hyperparameters as \ours\textsubscript{1}, but with updated annotations. this iteration of labeling and training gives \ours\textsubscript{2}. in Tab.~\ref{tab:zoo3d012}, we overview data involved in the training process of \ours\textsubscript{0}, \ours\textsubscript{1}, \ours\textsubscript{2}. 

\section{Open-vocabulary 3D Object Detection from Multi-view Images}

\subsection{Posed Images}

In the pose-aware scenario, our method accepts a set of images $\mathcal{I}$ along with camera extrinsics $\mathcal{R}$ and camera intrinsic $K$. In real applications, camera poses may be obtained from IMU or integrated tracking software.

Since we are able to handle point clouds, all we need is to transform images into a point cloud. For this purpose, we employ DUSt3R~\cite{wang2024dust3r}. It can either infer camera poses or take them as auxiliary inputs, making it a universal solution for both pose-aware and pose-agnostic inference. Last but not least, opting for DUSt3R allows keeping the whole pipeline zero-shot and prevent data leakage, since contrary to some latest methods~\cite{wang2025vggt} it was not trained on ScanNet.

For input images, DUSt3R returns dense depth maps, that are fused into a TSDF volume using ground truth camera poses. The final step of this pipeline is the point cloud extraction, and the task is thus reduced to the point cloud-based 3D object detection.

\subsection{Unposed Images}

In the third scenario, our model is given images $\mathcal{I}$ without known camera intrinsic $K$ and extrinsics $\mathcal{R}$, which is the typical scenario for smartphone applications or capturing done with consumer cameras.

\begin{figure}
\centering \small
\setlength{\tabcolsep}{0pt}
\resizebox{\linewidth}{!}{
\begin{tabular}{ccccc}
\multirow{2}{*}{Ground Truth} & \multicolumn{3}{c}{\ours\textsubscript{1} predictions from} & \multirow{2}{*}{Text prompt} \\
& Point cloud & Posed RGB & Unposed RGB \\
\includegraphics[width=0.14\textwidth, valign=c]{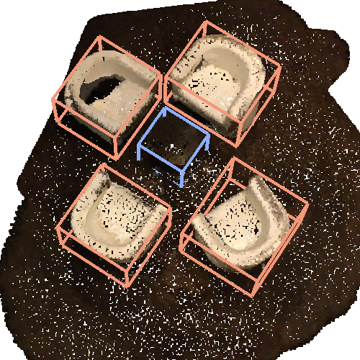} & 
\includegraphics[width=0.14\textwidth, valign=c]{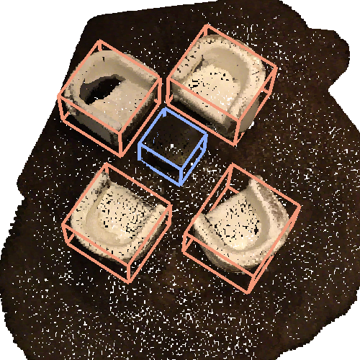} & 
\includegraphics[width=0.14\textwidth, valign=c]{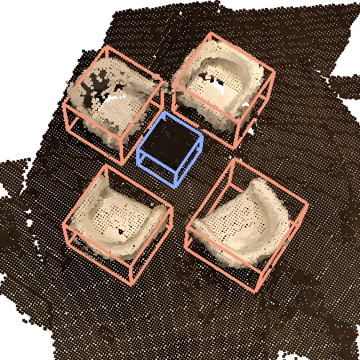} & \includegraphics[width=0.14\textwidth, valign=c]{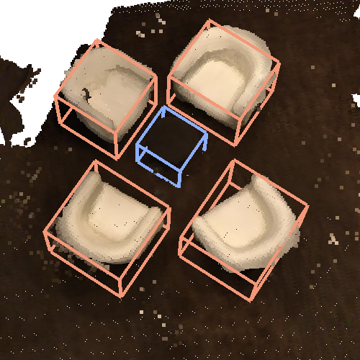} & \makecell[l]{\textit{a photo of} \\ \textcolor[rgb]{0.96,0.6,0.48}{\textit{armchair}} \\ \textcolor[rgb]{0.55,0.69,1.}{\textit{table}}} \\
\includegraphics[width=0.14\textwidth, valign=c]{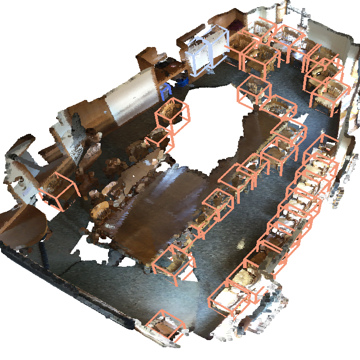} & 
\includegraphics[width=0.14\textwidth, valign=c]{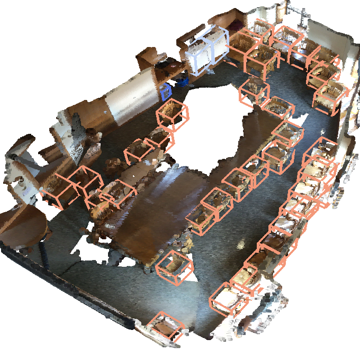} & 
\includegraphics[width=0.14\textwidth, valign=c]{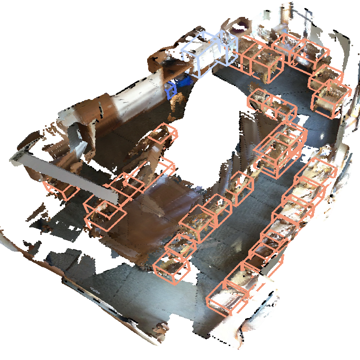} & \includegraphics[width=0.14\textwidth, valign=c]{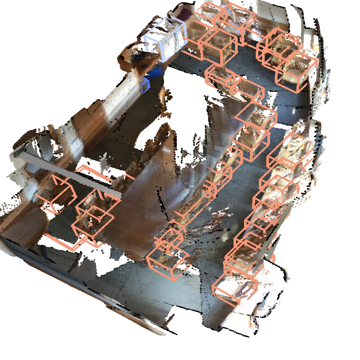} & \makecell[l]{\textit{a photo of} \\ \textcolor[rgb]{0.96,0.6,0.48}{\textit{chair}} \\ \textcolor[rgb]{0.75 0.83 0.96}{\textit{whiteboard}} \\ \textcolor[rgb]{0.35,0.47,0.89}{\textit{trash can}}} \\
\includegraphics[width=0.14\textwidth, valign=c]{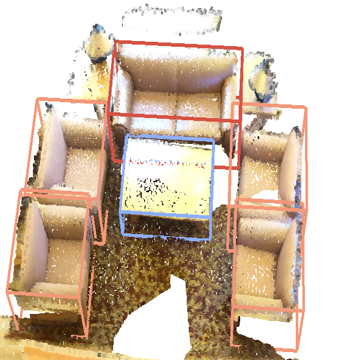} & 
\includegraphics[width=0.14\textwidth, valign=c]{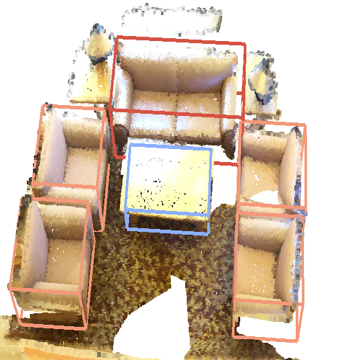} & 
\includegraphics[width=0.14\textwidth, valign=c]{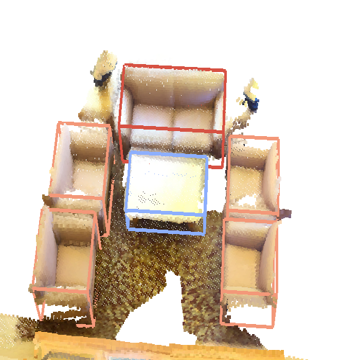} & \includegraphics[width=0.14\textwidth, valign=c]{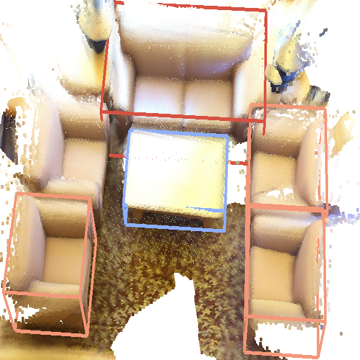} & \makecell[l]{\textit{a photo of} \\ \textcolor[rgb]{0.96,0.6,0.48}{\textit{armchair}} \\ \textcolor[rgb]{0.55,0.69,1.}{\textit{table}} \\ \textcolor[rgb]{0.84,0.32,0.26}{\textit{couch}}} \\
\end{tabular}
}
\caption{Qualitative results of \ours\textsubscript{1} on ScanNet200.}
\label{fig:qualitative}
\end{figure}

DUSt3R shines in this most challenging setting, inferring depth maps and camera poses within a single end-to-end framework. Same as in the previous case, the predicted depth maps are fused, and the point cloud is extracted from the TSDF volume.

\section{Experiments}

\paragraph{Datasets}

ScanNet~\cite{dai2017scannet} is a widely used dataset containing 1201 training and 312 validation scans. Following~\cite{qi2019votenet}, we estimate axis-aligned 3D bounding boxes from semantic per-point labels. We report results obtained in four benchmarks based on ScanNet: ScanNet10~\cite{lu2022ov-3detic}, ScanNet20~\cite{lu2023ov-3det}, ScanNet60~\cite{wang2024ov-uni3detr} and ScanNet200~\cite{rozenberszki2022scannet200}, containing 10, 20, 60, and 200 classes, respectively. In the ScanNet60 benchmark, models are trained with ground truth 3D bounding boxes of objects from the first 10 classes in a class-agnostic regime, and evaluated on all 60 classes in the open-vocabulary mode. In other benchmarks, annotated 3D bounding boxes are not seen during training. For all ScanNet-based benchmarks, we don't use ground truth 3D bounding boxes and report mean average precision (mAP) under IoU thresholds of 0.25 and 0.5.

ARKitScenes~\cite{baruch2021arkitscenes} is an RGB-D dataset of 4493 training scans and 549 validation scans. These scans contain RGB-D frames along with 3D object bounding box annotations of 17 categories. We use ARKitScenes for class-agnostic evaluation, following the protocol proposed by OpenM3D~\cite{hsu2025openm3d}, and report standard precision and recall.

\paragraph{Implementation details}

Zero-shot \ours\textsubscript{0} is training-free and uses the same hyperparameters as the original MaskClustering~\cite{yan2024maskclustering} during the inference. Self-supervised \ours\textsubscript{1} inherits the training hyperparameters of TR3D~\cite{rukhovich2023tr3d}. To control the size of input scenes, we sample a maximum of 100,000 points per scene during both training and inference. During inference, candidate detections are filtered using NMS with an IoU threshold of 0.5. Architecture-wise, we modify TR3D as described in Sec.~\ref{ssec:self-supervised}, keeping the rest of architecture the same as in the original TR3D. Our open-vocabulary module uses CLIP ViT-H/14~\cite{radford2021learning} and SAM 2.1 (Hiera-L)~\cite{ravi2024sam2}. For each scene, we sample 45 images uniformly. Images are resized into 480 $\times$ 640 in all experiments.

\subsection{Comparison with Prior Work}

\paragraph{Open-vocabulary 3D object detection from point clouds}

We benchmark the proposed approach in the most common scenario with ground truth point clouds as inputs and report scores on multiple benchmark on ScanNet with 10, 20, 60, and 200 classes. As reported in Tab.~\ref{tab:results-scannet}, our method demonstrates significant performance gains w.r.t. prior state-of-the-art open-vocabulary 3D object detection approaches ($+11.9$ mAP\textsubscript{25} on ScanNet20, $+12.6$ mAP\textsubscript{25} on ScanNet60 w.r.t. OV-Uni3DETR).
Overall, with consistent improvement over all existing methods, \ours{} sets a new state-of-the-art in 3D open-vocabulary object detection. According to Tab.~\ref{tab:results-scannet10} presenting the results on the ScanNet10 benchmark, we achieve competitive results even in a zero-shot setting, while our self-supervised model outperforms the closest competitor, OV-Uni3Detr, by +10.4 mAP25. 

\begin{table}[t!]
\centering 
\begin{tabular}{lcc}
\toprule
Method & mAP\textsubscript{25} & mAP\textsubscript{50} \\
\midrule
OV-3DETIC\textsuperscript{\dag}~\cite{lu2022ov-3detic}  & 12.7 & - \\
FM-OV3D~\cite{zhang2024fm-ov3d} & 17.0 & - \\
FM-OV3D\textsuperscript{\dag}~\cite{zhang2024fm-ov3d} & 21.5 & - \\
Object2Scene\textsuperscript{\dag}~\cite{zhu2023object2scene}  & 24.6 & -  \\
INHA\textsuperscript{\dag}~\cite{jiao2024inha} & 30.1 & -  \\
GLIS \cite{peng2024glis} & 30.9 & -  \\
OV-Uni3DETR\textsuperscript{\dag} \cite{wang2024ov-uni3detr}  & 34.1 & -  \\
\rowcolor{blue!5} \ours\textsubscript{0} & 42.1 & 28.7 \\
\rowcolor{blue!5} \ours\textsubscript{1} & \textbf{44.5} & \textbf{31.5} \\
\bottomrule
\end{tabular}
\caption{Results of open-vocabulary 3D object detection from point clouds on the ScanNet10 benchmark. Methods with \textsuperscript{\dag} utilize 3D bounding box annotations during training.}
\label{tab:results-scannet10}
\end{table}

\begin{table*}
\centering
\begin{tabular}{llccccccc}
\toprule
& \multirow{2}{*}{Method}  & \multirow{2}{*}{Zero-shot} & \multicolumn{2}{c}{ScanNet20}  & \multicolumn{2}{c}{ScanNet60} & \multicolumn{2}{c}{ScanNet200} \\
& & & mAP\textsubscript{25} & mAP\textsubscript{50} & mAP\textsubscript{25} & mAP\textsubscript{50} & mAP\textsubscript{25} & mAP\textsubscript{50} \\
\midrule
\multicolumn{7}{l}{\textit{Point cloud + posed images}} \\
\rowcolor{blue!5} & \ours\textsubscript{0} & \cmark & 36.0 & 21.9 & 29.6 & 16.5 & 32.1 & 17.6 \\
\rowcolor{blue!5} & \ours\textsubscript{1} & \xmark & \textbf{46.1} & \textbf{30.9} & \textbf{48.5} & \textbf{29.6} & \textbf{51.0} & \textbf{31.0} \\
\midrule
\multicolumn{7}{l}{\textit{Posed images}} \\
& OV-3DET \cite{lu2023ov-3det} $\rightarrow$ OpenM3D \cite{hsu2025openm3d} & \xmark  & -  & -  & - & - & 19.5 & - \\
& SAM3D \cite{yang2023sam3d} $\rightarrow$ OpenM3D \cite{hsu2025openm3d} & \xmark  & -  &  - & - & - & 23.8 & - \\
& OpenM3D \cite{hsu2025openm3d} & \xmark  & -  & - & - & - & 26.9 & - \\
\rowcolor{blue!5} & DUSt3R \cite{wang2024dust3r} $\rightarrow$ \ours\textsubscript{0}  & \cmark & 31.3 & 12.4 & 22.0 & 7.3 & 22.4 & 6.8 \\
\rowcolor{blue!5} & DUSt3R \cite{wang2024dust3r} $\rightarrow$ \ours\textsubscript{1} & \xmark & \textbf{40.1} & \textbf{18.9} & \textbf{35.0} & \textbf{14.0} & \textbf{36.1} & \textbf{13.9} \\
\midrule
\multicolumn{7}{l}{\textit{Unposed images}} \\
\rowcolor{blue!5} & DUSt3R \cite{wang2024dust3r} $\rightarrow$ \ours\textsubscript{0} & \cmark & 16.4 & 2.5 & 10.1 & 1.5 & 10.3 & 1.5 \\
\rowcolor{blue!5} & DUSt3R \cite{wang2024dust3r} $\rightarrow$ \ours\textsubscript{1} & \xmark & \textbf{22.7} & \textbf{5.7} & \textbf{18.3} & \textbf{4.0} & \textbf{19.0} & \textbf{3.9} \\
\bottomrule
\end{tabular}
\caption{Results of class-agnostic 3D object detection on the ScanNet dataset across three benchmarks (with 20, 60, and 200 classes), and three input modalities (points cloud, posed and unposed multi-view images).}
\label{tab:results-class-agnostic}
\end{table*}

\paragraph{Open-vocabulary 3D object detection from posed images} is a more challenging task gaining attention recently, so in this competitive track we test against approaches leveraging state-of-the-art techniques. Still, \ours{} shines, with +10.1  mAP\textsubscript{25}  ScanNet200 w.r.t. OpenM3D and +12.7 mAP\textsubscript{25} ScanNet60 w.r.t. OV-Uni3DETR achieved in self-supervised mode. Surprisingly, even without access to ground truth point clouds, \ours{} outperforms point cloud-based approaches in both zero-shot and self-supervised scenario. This is a strong evidence that with a thoughtfully designed pipeline, we can beat state-of-the-art without training and having less data on inference.

\paragraph{Open-vocabulary 3D object detection from unposed images} In the third track, there are no predecessors reporting 3D open-vocabulary object detection results. Respectively, we also obtain reference numbers with a combination of DUSt3R $\rightarrow$ \ours{}\textsubscript{0} and DUSt3R $\rightarrow$ \ours{}\textsubscript{1}. As could be expected, DUSt3R $\rightarrow$ \ours{}\textsubscript{1} outperforms DUSt3R $\rightarrow$ \ours{}\textsubscript{0}. What is more exciting is that the performance of our self-supervised \ours{}\textsubscript{1} is close to the one of OV-Uni3DETR using point clouds. In other words, being exposed to the same training data, but with neither point clouds nor even camera poses on inference, our method can deliver competitive accuracy to point cloud-based solutions.

\paragraph{Class-agnostic 3D object detection} is less studied, with existing methods only using posed images as inputs and reporting quality on ScanNet200. Still, we report metrics across all three input modalities and three benchmarks to establish the baseline for future research. Our zero-shot model is inferior to OpenM3D, while self-supervised \ours{}\textsubscript{1} surpasses it by a large margin, achieving +9 mAP\textsubscript{25}. 

\subsection{Ablation Experiments}

\paragraph{Class-agnostic 3D object detection (OpenM3D protocol)} implies calculating precision and recall for class-agnostic 3D bounding boxes on ScanNet200 and ARKitScenes. This alternative evaluation protocol is more coherent with how those boxes are used in our pipeline: we omit low-confidence boxes and only pass reliable detections to the open-vocabulary module, so the predictions are turned binary rather than probabilistic. As seen from Tab.~\ref{tab:ablation-precision-recall}, \ours\textsubscript{0} only slightly outperforms previous state-of-the-art in precision and recall calculated at the threshold of 0.25. When limiting the evaluation scope to more accurate detections with a threshold of 0.5, the superiority of \ours\textsubscript{0} becomes more prominent, i.e., metrics on ARKitScenes double, while the ScanNet scores rise by approx 50\%.


\begin{table}
\centering
\resizebox{\linewidth}{!}{
\begin{tabular}{lcccccccc}
\toprule
\multirow{2}{*}{Method} & \multicolumn{4}{c}{ScanNet200} &  \multicolumn{4}{c}{ARKitScenes} \\
\cmidrule(r{0.2em}){2-5} \cmidrule(l{0.2em}){6-9} 
&  P\textsubscript{25} & R\textsubscript{25} & P\textsubscript{50} & R\textsubscript{50} & P\textsubscript{25} & R\textsubscript{25} & P\textsubscript{50} & R\textsubscript{50}   \\
\midrule
OV-3DET \cite{lu2023ov-3det} & 11.6 & 21.3 & 4.4 & 8.0 & 3.7 & 32.4 & 0.9 & 7.9 \\
SAM3D \cite{yang2023sam3d} & 14.5 & 57.7 & 9.0 & 36.1 & 6.0 & 43.8 & 1.5 & 10.9 \\
OpenM3D  \cite{hsu2025openm3d} & 32.0 & 58.3 & 18.1 & 33.0 & 5.9 & 51.9 & 1.6 & 13.7 \\
\rowcolor{blue!5} \ours\textsubscript{0} & \textbf{35.9} & \textbf{58.7} & \textbf{27.9} & \textbf{45.6} & \textbf{8.4} & \textbf{52.0} & \textbf{4.3} & \textbf{28.2} \\
\bottomrule
\end{tabular}}
\caption{Results of class-agnostic 3D object detection on ScanNet200 and ARKitScenes, OpenM3D evaluation protocol. P -- precision, R -- recall.}
\label{tab:ablation-precision-recall}
\end{table}

\paragraph{Inference time}

Up to a certain degree, the superior quality of our approach should be attributed to the significantly longer processing time. In Tab.~\ref{tab:ablation-time}, we report inference time of \ours\textsubscript{0} and \ours\textsubscript{1} in comparison to the OpenM3D, an efficient single-stage feed-forward method. OpenM3D does not require a separate open-vocabulary module with CLIP, achieving impressive performance that comes at the cost of accuracy. Still, the most time-expensive component of our pipeline is point cloud reconstruction, which is performed with DUSt3R.
The latency of \ours\textsubscript{0} and \ours\textsubscript{1} also differs due to distinct class-agnostic 3D object detection strategies applied. While \ours\textsubscript{0} builds a mask graph, which can be slow, the self-supervised method uses a lightweight sparse-convolutional model. However, mask clustering procedure assigns instance labels unambiguously by design, while detection model produces numerous duplicates. Respectively, more detections should be processed at the open-vocabulary stage, leading to 7x increase of inference time. 

\begin{table}
\centering
\resizebox{\linewidth}{!}{
\begin{tabular}{lccc}
\toprule
Method & Reconst. & Detect. & Open-vocab. \\
\midrule
OpenM3D \cite{hsu2025openm3d} & -- & 0.3 & 0.01 \\
DUSt3R \cite{wang2024dust3r} $\rightarrow$ \ours\textsubscript{0} & 294.3 & 56.3 & 12.6 \\
DUSt3R \cite{wang2024dust3r} $\rightarrow$ \ours\textsubscript{1} & 294.3 & 0.04 & 84.0 \\
\bottomrule
\end{tabular}
}
\caption{Inference time (sec) of 3D object detection from posed images on ScanNet200, component-wise: reconstruction (DUSt3R), class-agnostic 3D object detection, and open-vocabulary module.}
\label{tab:ablation-time}
\end{table}

\begin{table}
\centering
\begin{tabular}{cccccc}
\toprule
\multirow{2}{*}{\# Images} & \multicolumn{2}{c}{Posed} & \multicolumn{2}{c}{Unposed}  \\
& mAP\textsubscript{25} & mAP\textsubscript{50} & mAP\textsubscript{25} & mAP\textsubscript{50} \\
\midrule
15 & 9.6 & 3.9 & 5.8 & 1.9 \\
25 & 12.5 & 5.2 & 7.1 & 2.6 \\
35 & 14.0 & 5.7 & 8.1 & 2.7 \\
\textbf{45} & \textbf{14.3} & \textbf{6.2} & \textbf{8.3} & \textbf{2.9} \\
\bottomrule
\end{tabular}
\caption{Results of open-vocabulary 3D object detection from posed and unposed images on ScanNet200 with varying number of images.}
\label{tab:ablation-n-frames}
\end{table}

\begin{table}
\centering
\begin{tabular}{lcc}
\toprule
Method & mAP\textsubscript{25} & mAP\textsubscript{50} \\
\midrule
DROID-SLAM \cite{teed2021droid} $\rightarrow$ \ours{}\textsubscript{1}  & 2.4 & 0.6  \\
DUSt3R \cite{wang2024dust3r} $\rightarrow$ \ours{}\textsubscript{1}  & \textbf{19.0} & \textbf{3.9} \\
\bottomrule
\end{tabular}\caption{Results of self-supervised method from unposed images in class-agnostic mode on ScanNet200 with different pose estimation methods.}
\label{tab:ablation-slam}
\end{table}

\paragraph{Number of images per scene}

In Tab.~\ref{tab:ablation-n-frames}, we ablate \ours\textsubscript{0} against varying number of images per scenes. The model hits the peak with 45 frames, and performs on par with OpenM3D and OV-Uni3DETR using 20 frames. On ScanNet200, \ours\textsubscript{0} outperforms OpenM3D with as few as 15 images.

\paragraph{DUSt3R vs DROID-SLAM}

As an alternative to DUSt3R for point cloud reconstruction, we use DROID-SLAM, a well-known reconstruction method based on entirely different working principles. According to Tab.~\ref{tab:ablation-slam}, quality drops dramatically when switching to DROID-SLAM: apparently, DROID-SLAM fails to produce reconstructions of quality that allows localizing and recognizing 3D objects reliably.

\paragraph{Iterative training}

is ablated in Tab.~\ref{tab:ablation-iteration}. \ours\textsubscript{2} denotes the self-supervised model trained with annotations generated by \ours\textsubscript{1}. Evidently, the iterative training process results in an improvement between steps 0 and 1, and between 1 and 2. However, performance gradually saturates, and by the third iteration, the metrics cease to improve.

\paragraph{Open-vocabulary module}

consists of three components, namely, occlusion filter, SAM-based mask refinement and multi-scale processing. We evaluate the contribution of each component by switching if off and measuring quality degradation. Instead of SAM mask refinement, we simply project all the object's points onto the image and approximate mask with a bounding-box tightly enclosing the projected points. As can be observed in Tab.~\ref{tab:ablation-ov}, multi-scale processing improves mAP\textsubscript{50}, while SAM mask refinement contributes to mAP\textsubscript{25}.

\begin{table}
\centering
\begin{tabular}{lccc}
\toprule
Method & mAP\textsubscript{25} & mAP\textsubscript{50} \\
\midrule
DUSt3R \cite{wang2024dust3r} $\rightarrow$ \ours\textsubscript{0} & 22.4 & 6.8 \\
DUSt3R \cite{wang2024dust3r} $\rightarrow$ \ours\textsubscript{1} & 36.1 & 13.9 \\
DUSt3R \cite{wang2024dust3r} $\rightarrow$ \ours\textsubscript{2} & \textbf{37.6} & \textbf{14.8} \\
\bottomrule
\end{tabular}
\caption{Results of class-agnostic 3D object detection from posed images on ScanNet200 with iterative training.}
\label{tab:ablation-iteration}
\end{table}

\begin{table}
\centering
\begin{tabular}{lccc}
\toprule
Module & mAP\textsubscript{25} & mAP\textsubscript{50} \\
\midrule
base & 14.7 & 5.7 \\
+ occlusion filter & 14.8 & 5.7 \\
+ SAM mask refinement & 15.4 & 5.7 \\
+ multi-scale & \textbf{16.5} & \textbf{6.3} \\
\bottomrule
\end{tabular}
\caption{Results of open-vocabulary 3D object detection from posed images on ScanNet200, with different configurations of the open-vocabulary module.}
\label{tab:ablation-ov}
\end{table}

\section{Conclusion}

We proposed a first-in-class zero-shot \ours\textsubscript{0} and a self-supervised \ours\textsubscript{1} open-vocabulary 3D object detection methods. Besides, we adapted \ours{} to work directly with posed and even unposed images, so that point clouds are not required. Across multiple benchmarks, both \ours\textsubscript{0} and \ours\textsubscript{1} achieve state-of-the-art results in open-vocabulary 3D detection. As a possible future research direction, we consider speeding up the pipeline with elaborate procedures of harvesting spatial information from 2D foundation models: a faster reconstruction method, a better segmentation model, a more efficient open-vocabulary label assignment, and extensive use of large language models.

\appendix

In Appendix, we provide additional quantitative scores, including evaluation results on ScanNet++~\cite{yeshwanth2023scannet++} and ARKitScenes~\cite{baruch2021arkitscenes}, in Sec.~\ref{sec:supp-quantitative}, report results of ablation experiments in Sec.~\ref{sec:supp-ablation}, and show more visualizations in Sec.~\ref{sec:supp-visualization}.

\section{Quantitative Results} \label{sec:supp-quantitative}

\paragraph{ScanNet++} As can be seen from Tab.~\ref{tab:results-scannet++}, \ours\textsubscript{0} sets state-of-the-art in ScanNet++~\cite{yeshwanth2023scannet++} even compared with fully-supervised methods. Despite not being exposed to any annotations or even training scenes, it outperforms models that utilize both scans and labeled 3D bounding boxes during the training. 

\paragraph{ScanNet60} results are reported in Tab.~\ref{tab:results-scannet60}. For objects of \textit{base} categories, \ours{} falls beyond training-based competitors, which could be expected, since neither of our methods has access to ground truth 3D bounding boxes. For \textit{novel} objects not seen during training, \ours{} scores first in the leaderboard.

\begin{table}[b!]
\centering
\begin{tabular}{lcc}
\toprule
Method & mAP\textsubscript{25} & mAP\textsubscript{50} \\
\midrule
TR3D\textsuperscript{\dag}~\cite{rukhovich2023tr3d} & 26.2 & 14.5  \\
UniDet3D\textsuperscript{\dag}~\cite{kolodiazhnyi2025unidet3d} & 26.4 & 17.2 \\
\rowcolor{blue!5} \ours\textsubscript{0} & \textbf{26.5} & \textbf{18.3} \\
\bottomrule
\end{tabular}
\caption{3D object detection results from points clouds on ScanNet++ dataset. \textsuperscript{\dag} is for fully-supervised method utilized labeled 3D bounding boxes during training.}
\label{tab:results-scannet++}
\end{table}

\begin{table}[h!]
\resizebox{\linewidth}{!}{
\centering
\begin{tabular}{llcccc}
\toprule
& \multirow{2}{*}{Method} & Zero- & \multirow{2}{*}{Novel} & \multirow{2}{*}{Base} & \multirow{2}{*}{All}  \\
& & shot \\
\midrule
& Det-PointCLIPv2\textsuperscript{\dag}~\cite{zhu2023pointclip} & \xmark & 0.1 & 1.0 & 0.2 \\
& 3D-CLIP\textsuperscript{\dag}~\cite{radford2021learning}  & \xmark & 2.5 & 11.2 & 3.9 \\
& CoDA\textsuperscript{\dag}~\cite{cao2023coda} & \xmark & 6.5 & 21.6 & 9.0  \\
& INHA\textsuperscript{\dag}~\cite{jiao2024inha} & \xmark & 7.8 & 25.1 & 10.7 \\
& OV-Uni3DETR\textsuperscript{\dag}~\cite{wang2024ov-uni3detr} & \xmark & 13.7 & \textbf{48.1} & 19.4 \\
\rowcolor{blue!5} & \ours\textsubscript{0} & \cmark & 29.3 & 16.2 & 27.1  \\
\rowcolor{blue!5} & \ours\textsubscript{1} & \xmark & \textbf{33.6} & 24.4 & \textbf{32.0} \\
\bottomrule
\end{tabular}
}
\caption{Results of open-vocabulary 3D object detection of \textit{base}, \textit{novel}, and \textit{all} object categories on ScanNet60. Methods marked with \textsuperscript{\dag} use ground truth 3D bounding boxes for objects of \textit{base} classes.}
\label{tab:results-scannet60}
\end{table}

\paragraph{3D segmentation baselines}

Open-vocabulary 3D object detection metrics on ScanNet are listed in Tab.~\ref{tab:ablation-zshseg}. To establish baselines, we adapt state-of-the-art 3D instance segmentation approaches by simply enclosing each predicted mask with a 3D bounding box. Obviously, while \ours\textsubscript{0} is built upon MaskClustering, our open-vocabulary assignment strategy is way more effective compared to the one used in the original MaskClustering. 

\begin{table}
\centering
\begin{tabular}{lccc}
\toprule
Method & Venue & mAP\textsubscript{25} & mAP\textsubscript{50} \\
\midrule
MaskClustering~\cite{yan2024maskclustering} & CVPR'24 & 13.4 & 8.5 \\
OnlineAnySeg~\cite{tang2025onlineanyseg} & CVPR'25 & 19.0 & 12.3 \\
\rowcolor{blue!5}   \ours\textsubscript{0} & - & \textbf{21.1} & \textbf{14.1} \\
\bottomrule
\end{tabular}
\caption{Comparison with zero-shot 3D instance segmentation methods on ScanNet200.}
\label{tab:ablation-zshseg}
\end{table}

\paragraph{ARKitScenes} was used to train DUSt3R, so we cannot use it in zero-shot experiments for fair comparison. In this series of experiments, DUSt3R is replaced with VGGT to preserve methodological purity. Since VGGT does not support camera poses as inputs natively, we only report quality in point cloud-based and unposed images-based tracks in Tab.~\ref{tab:results-arkitscenes}.

\begin{table*}
\resizebox{0.95\textwidth}{!}{
\begin{tabular}{c|c|cccccccccc}
\toprule 
Methods & mAP\textsubscript{25} & toilet & bed  & chair & sofa & dresser & table & cabinet & bookshelf & pillow & sink \\
\midrule
OV-3DET~\cite{lu2023ov-3det} & 18.0 & 57.3 & 42.3 & 27.1 & 31.5 & 8.2 & 14.2 & 3.0 & 5.6 & 23.0 & 31.6  \\
CoDA~\cite{cao2023coda} & 19.3 & 68.1 & 44.1 &  28.7 &  44.6 &  3.4 & 20.2 &  5.3 &  0.1 &  28.0 &  45.3     \\
OV-Uni3DETR~\cite{wang2024ov-uni3detr} & 25.3 & 86.1 & 50.5 & 28.1 & 31.5 & 18.2 & \textbf{24.0} & 6.6 & 12.2 & 29.6 & \textbf{54.6} \\
\rowcolor{blue!5} \ours\textsubscript{0} & 34.7 & \textbf{91.1} & 51.2 & 53.4 & 60.5 & 31.9 & 20.2 & 12.9 & \textbf{41.9} & 32.2 & 25.7 \\
\rowcolor{blue!5} \ours\textsubscript{1} & \textbf{37.2} & 78.4 & \textbf{54.4} & \textbf{74.4} & \textbf{65.5} & \textbf{33.6} & 19.1 &\textbf{ 14.1} & 32.3 & \textbf{46.1} & 27.3 \\
\midrule
& & bathtub & refrigerator & desk & nightstand & counter & door &  curtain & box & lamp  & bag  \\
\midrule
 OV-3DET~\cite{lu2023ov-3det} & & 56.3 & 11.0 & 19.7 & 0.8 & 0.3 & 9.6 & 10.5 & 3.8 & 2.1 & 2.7   \\
CoDA~\cite{cao2023coda} & &  50.5 &  6.6 &  12.4 &  15.2 &  0.7 &  8.0 &  0.0  &  2.9 &  0.5 &  2.0 \\
OV-Uni3DETR~\cite{wang2024ov-uni3detr}  &  & 63.7 & 14.4 & \textbf{30.5} & 2.9 & \textbf{1.0} & 1.0 & \textbf{19.9} & 12.7 & 5.6 & \textbf{13.5} \\
\rowcolor{blue!5} \ours\textsubscript{0} &  & 50.0 & 50.5 & 11.2 & \textbf{59.2} & 0.1 & 21.1 & 18.2 & 17.8 & 34.8 & 9.8 \\
\rowcolor{blue!5} \ours\textsubscript{1} &  & \textbf{64.6} & \textbf{57.5} & 10.7 & 58.8 & 0.2 & \textbf{27.4} & 8.0 & \textbf{20.0} & \textbf{43.3} & 9.1 \\
\bottomrule
\end{tabular}
}
\caption{Per-class 3D object detection scores on the ScanNet20.}
\label{tab:per-category-scannet20}
\end{table*}

\begin{table}
\centering
\resizebox{\linewidth}{!}{
\begin{tabular}{llccc}
\toprule
& Method & Zero-shot & mAP\textsubscript{25} & mAP\textsubscript{50} \\
& & shot \\
\midrule
\multicolumn{4}{l}{\textit{Point cloud + posed images}} \\
& \ours\textsubscript{0} & \cmark & 24.4 & 11.0  \\
& \ours\textsubscript{1} & \xmark & \textbf{34.2} & \textbf{24.2} \\
\midrule
\multicolumn{4}{l}{\textit{Unposed images}} \\
& VGGT \cite{wang2025vggt} $\rightarrow$ \ours\textsubscript{0} & \cmark & 13.0 & 2.6 \\
& VGGT \cite{wang2025vggt} $\rightarrow$ \ours\textsubscript{1} & \xmark & \textbf{16.1}  & \textbf{3.5} \\
\bottomrule
\end{tabular}
}
\caption{Results of open-vocabulary 3D object detection from points clouds on ARKitScenes.}
\label{tab:results-arkitscenes}
\end{table}

\paragraph{Per-category scores} on ScanNet20 are given in Tab.~\ref{tab:per-category-scannet20}. Evidently, both our methods outperform the competitors by a large margin, with the most significant gains achieved for \textit{chair} (+45.7 mAP\textsubscript{25} for \ours\textsubscript{1} over previous state-of-the-art), \textit{sofa} (+20.9), \textit{bookshelf} (+19.7), \textit{refridgerator} (+43.1), \textit{nightstand} (+44.0), \textit{lamp} (+37.7). The average mAP\textsubscript{25} of \ours\textsubscript{0} is 9.4 higher than of any existing approach, while \ours\textsubscript{1} further expands the gap to 11.9 mAP\textsubscript{25}. 

\section{Ablation Experiments} \label{sec:supp-ablation}

\paragraph{VGGT vs DUSt3R for point cloud reconstruction} is evaluated in Tab.~\ref{tab:ablation-vggt}. Evidently, VGGT surpasses DUSt3R by a large margin, which can be expected, since VGGT was trained on ScanNet. Unfortunately, this also means that we cannot use it in the zero-shot setting; so despite the superior performance of VGGT, we employ DUSt3R as our primary method in our ScanNet-based experiments.

\paragraph{Level assignment strategy}

In Tab.~\ref{tab:ablation-assigner}, we ablate assigner parameters. In the class-agnostic mode, object classes remain unknown during the training, so we cannot apply the category-aware assignment scheme used in the original TR3D. Namely, we try assigning all objects to the 16 cm-level or 32 cm-level, or split the objects based on their size 50/50. The results demonstrate that assigning all objects to the 16-cm level yields the best performance.

\begin{table}[h!]
\centering
\resizebox{\linewidth}{!}{
\begin{tabular}{lccc}
\toprule
\multirow{2}{*}{Method}  & Trained on & \multirow{2}{*}{mAP\textsubscript{25}} & \multirow{2}{*}{mAP\textsubscript{50}} \\
& ScanNet & & \\
\midrule
DUSt3R \cite{wang2024dust3r} $\rightarrow$ \ours{}\textsubscript{1} & \xmark & 19.0 & 3.9 \\
VGGT \cite{wang2025vggt} $\rightarrow$ \ours{}\textsubscript{1} & \cmark & \textbf{28.2} & \textbf{6.4} \\
\bottomrule
\end{tabular}
}
\caption{Results of class-agnostic \ours\textsubscript{1} from unposed images on ScanNet200 with different pose estimation methods.}
\label{tab:ablation-vggt}
\end{table}

\begin{table}
\centering
\begin{tabular}{cccc}
\toprule
Object assignment & mAP\textsubscript{25} & mAP\textsubscript{50}  \\
\midrule
all objects to 32-cm level & 31.2 & 9.8 \\
50/50 & 34.5 & 12.3 \\
all objects to 16-cm level & \textbf{36.1} & \textbf{13.9} \\
\bottomrule
\end{tabular}
\caption{Results of class-agnostic \ours\textsubscript{1} from posed images on ScanNet200 with different assignment strategies.}
\label{tab:ablation-assigner}
\end{table}

\begin{table}[h!]
\centering
\begin{tabular}{lccc}
\toprule
 Alignment strategy & mAP\textsubscript{25} & mAP\textsubscript{50} \\
\midrule
first 2 poses & 4.1 & 0.8 \\
first depth + first pose & \textbf{8.3} & \textbf{2.9} \\
\bottomrule
\end{tabular}
\caption{Results of \ours\textsubscript{0} from unposed images on ScanNet200 with different alignment stratagies.}
\label{tab:ablation-alignment}
\end{table}

\paragraph{Alignment with g/t in image-based scenarios} 

In Tab.~\ref{tab:ablation-alignment}, we report results obtained with different alignment strategies. Here, point clouds are reconstructed from images; still, ground truth annotations are needed for evaluation, since scans must be transformed into common coordinate space before computing the metrics. To estimate an affine transformation that aligns ground truth and predicted point clouds, we apply two strategies. In the first strategy, we use both ground truth and predicted depth and pose for the first frame. The predicted pose is aligned with the ground truth pose, giving the rotation and translation. The scale is estimated from a single ground‑truth depth map as the median of per‑pixel ratios of ground‑truth to predicted depth.
In the second strategy, we use ground truth and predicted camera poses of two first frames in a sequence. The first predicted pose is aligned with the first ground truth pose, giving the rotation and translation. The relative scale coefficient is derived as a ratio of distances between two camera poses in predicted and ground truth scans.

\begin{figure*}[h!]
\centering \small
\resizebox{\linewidth}{!}{
\begin{tabular}{ccccc}
\multirow{2}{*}{Ground Truth} & \multicolumn{3}{c}{\ours\textsubscript{0} predictions from} & \multirow{2}{*}{Text prompt} \\
& Point cloud & Posed RGB & Unposed RGB \\
\includegraphics[width=0.20\linewidth, valign=c]{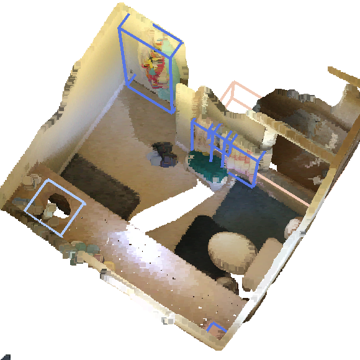} & 
\includegraphics[width=0.20\linewidth, valign=c]{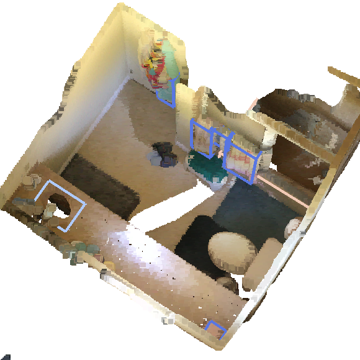} & 
\includegraphics[width=0.20\linewidth, valign=c]{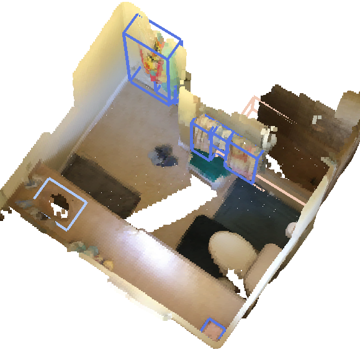} & \includegraphics[width=0.20\linewidth, valign=c]{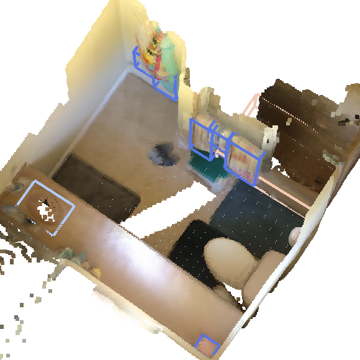} & \makecell[l]{\textit{a photo of} \\ \textcolor[rgb]{0.35,0.47,0.89}{\textit{towel}} \\ \textcolor[rgb]{0.69,0.79,0.99}{\textit{sink}} \\ \textcolor[rgb]{0.95,0.79,0.72}{\textit{bathtub}}} \\
\includegraphics[width=0.20\linewidth, valign=c]{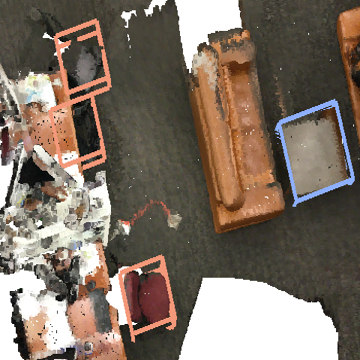} & 
\includegraphics[width=0.20\linewidth, valign=c]{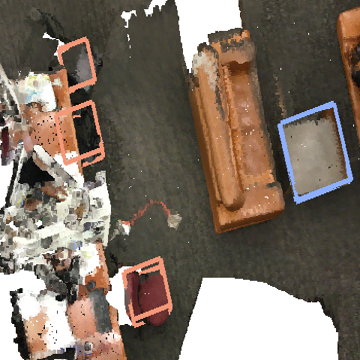} & 
\includegraphics[width=0.20\linewidth, valign=c]{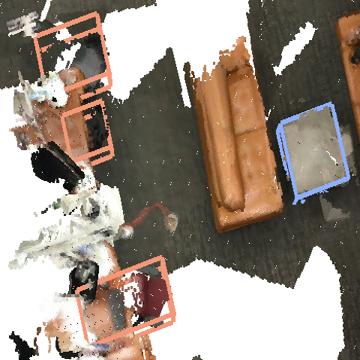} & \includegraphics[width=0.20\linewidth, valign=c]{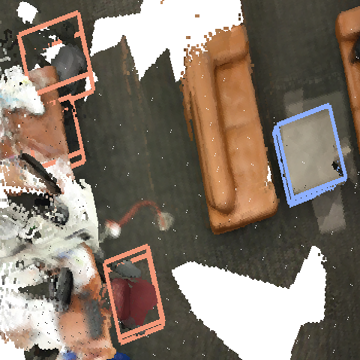} & \makecell[l]{\textit{a photo of} \\ \textcolor[rgb]{0.96,0.6,0.48}{\textit{chair}} \\ \textcolor[rgb]{0.55,0.69,1.}{\textit{coffee table}}} \\
\includegraphics[width=0.20\linewidth, valign=c]{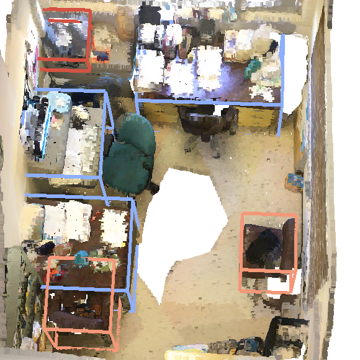} & 
\includegraphics[width=0.20\linewidth, valign=c]{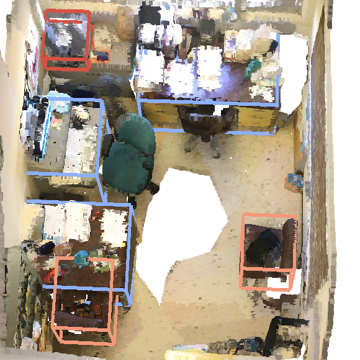} & 
\includegraphics[width=0.20\linewidth, valign=c]{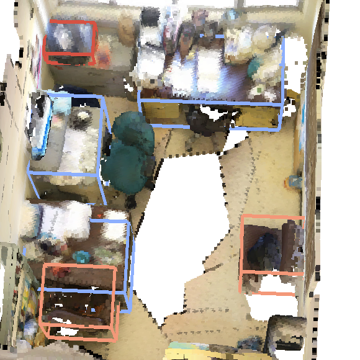} & \includegraphics[width=0.20\linewidth, valign=c]{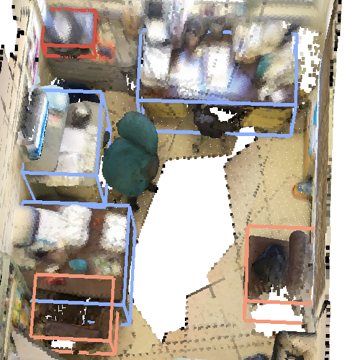} & \makecell[l]{\textit{a photo of} \\ \textcolor[rgb]{0.96,0.6,0.48}{\textit{chair}} \\ \textcolor[rgb]{0.55,0.69,1.}{\textit{desk}} \\ \textcolor[rgb]{0.84,0.32,0.26}{\textit{printer}}} \\
\end{tabular}
}
\caption{Qualitative results of \ours\textsubscript{0} on ScanNet200.}
\label{fig:qualitative-zoo3d0-scannet200}
\end{figure*}

\begin{figure}[h!]
\centering \small
\setlength{\tabcolsep}{0pt}
\resizebox{\linewidth}{!}{
\begin{tabular}{cccc}
\multirow{2}{*}{Ground Truth} & \multicolumn{2}{c}{\ours\textsubscript{0} predictions from} & \multirow{2}{*}{Text prompt} \\
& Point cloud & Unposed RGB \\
\includegraphics[width=0.25\linewidth, valign=c]
{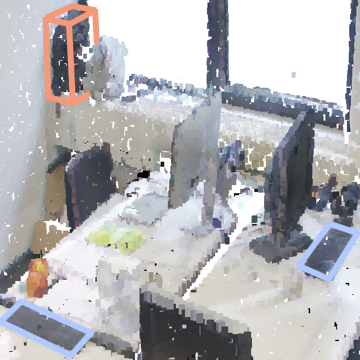} & 
\includegraphics[width=0.25\linewidth, valign=c]{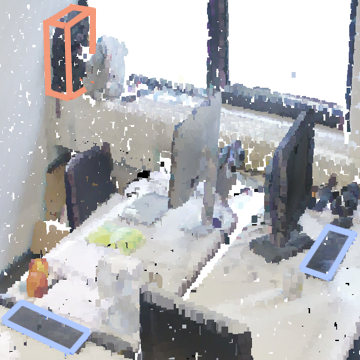} & 
\includegraphics[width=0.25\linewidth, valign=c]{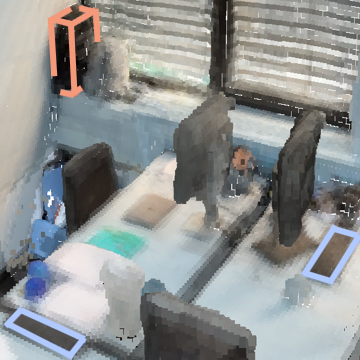} & \makecell[l]{\textit{a photo of} \\ \textcolor[rgb]{0.69,0.79,0.99}{\textit{keyboard}} \\ \textcolor[rgb]{0.96,0.6,0.48}{\textit{backpack}}} \\
\includegraphics[width=0.25\linewidth, valign=c]
{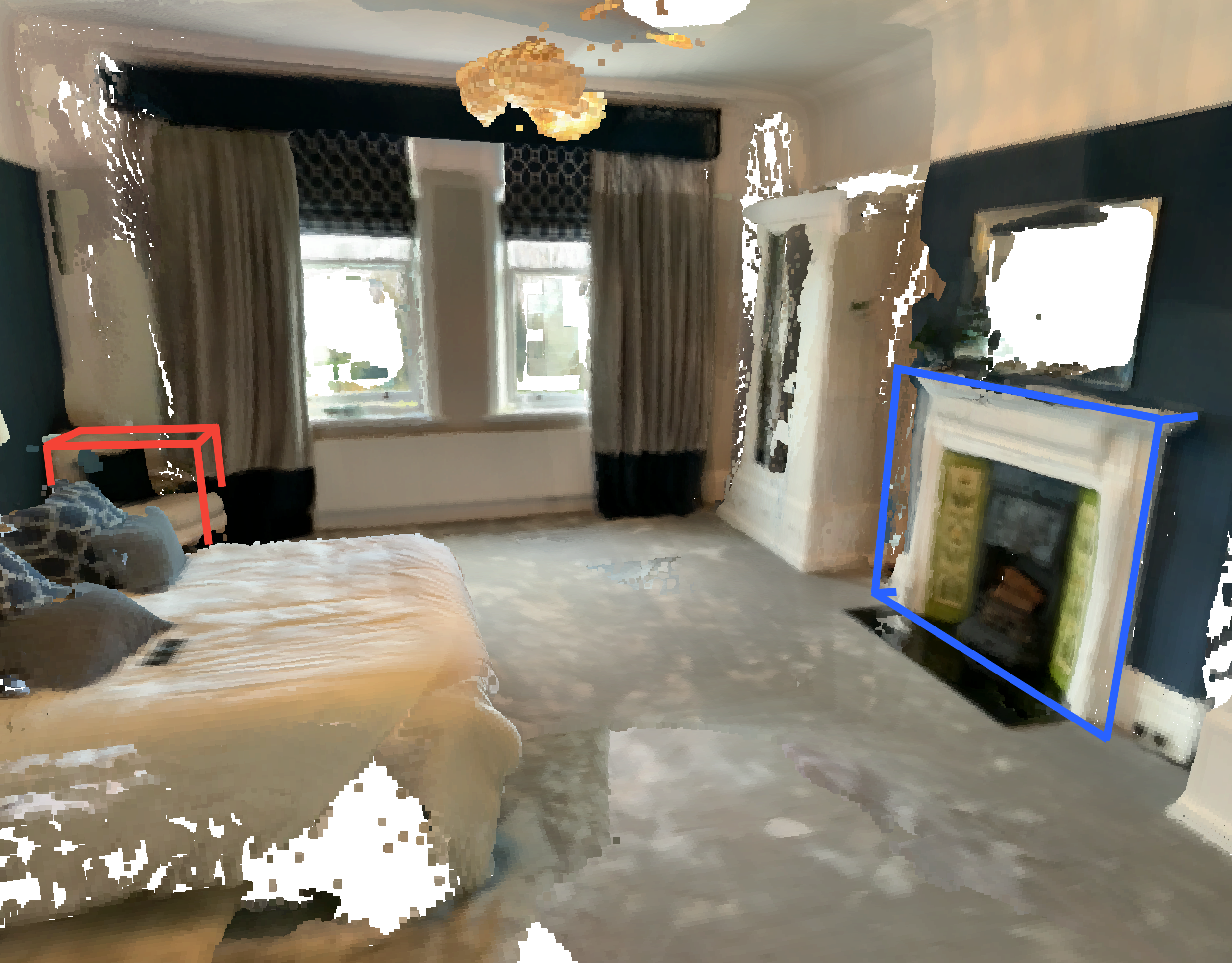} & 
\includegraphics[width=0.25\linewidth, valign=c]
{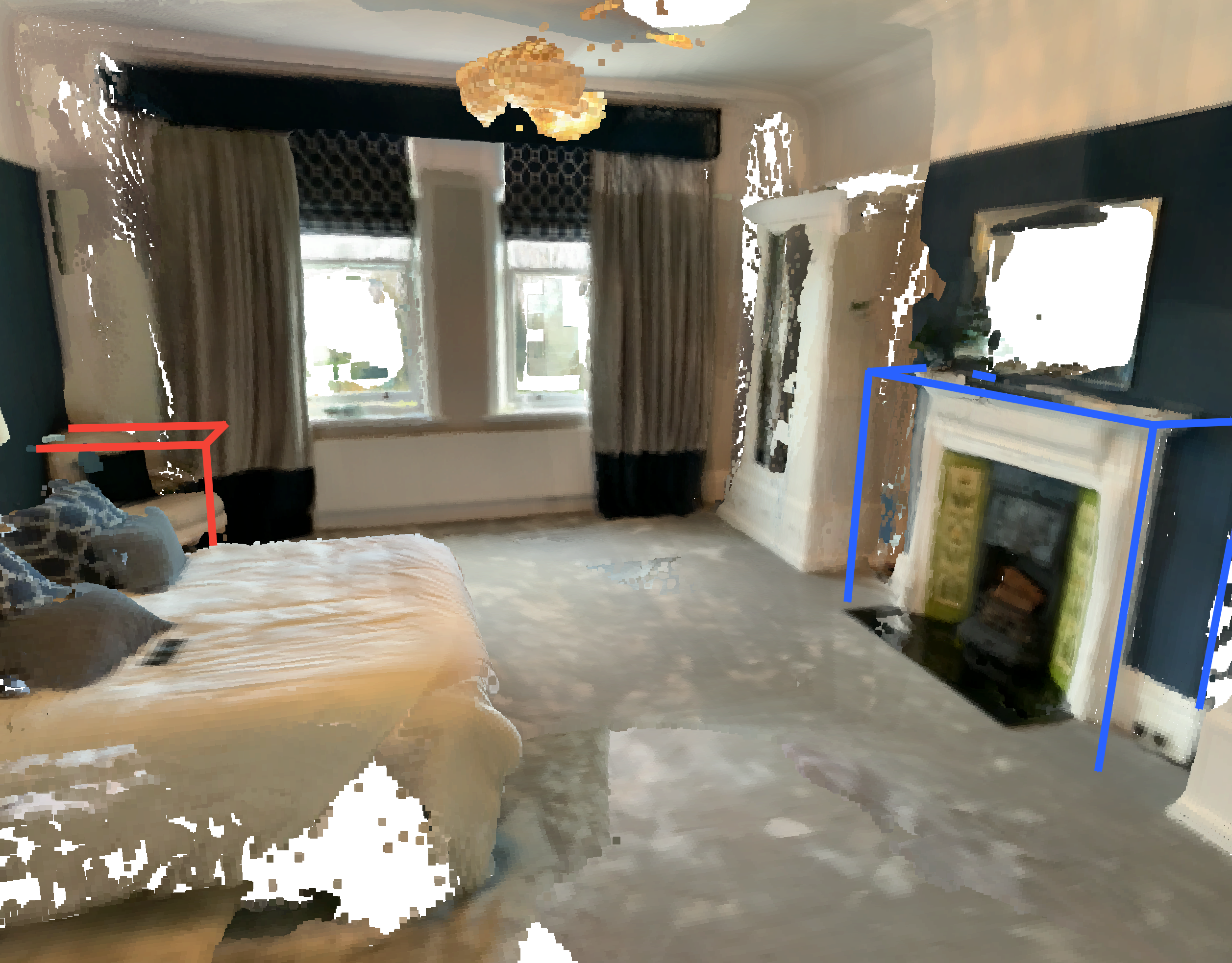} & 
\includegraphics[width=0.25\linewidth, valign=c]{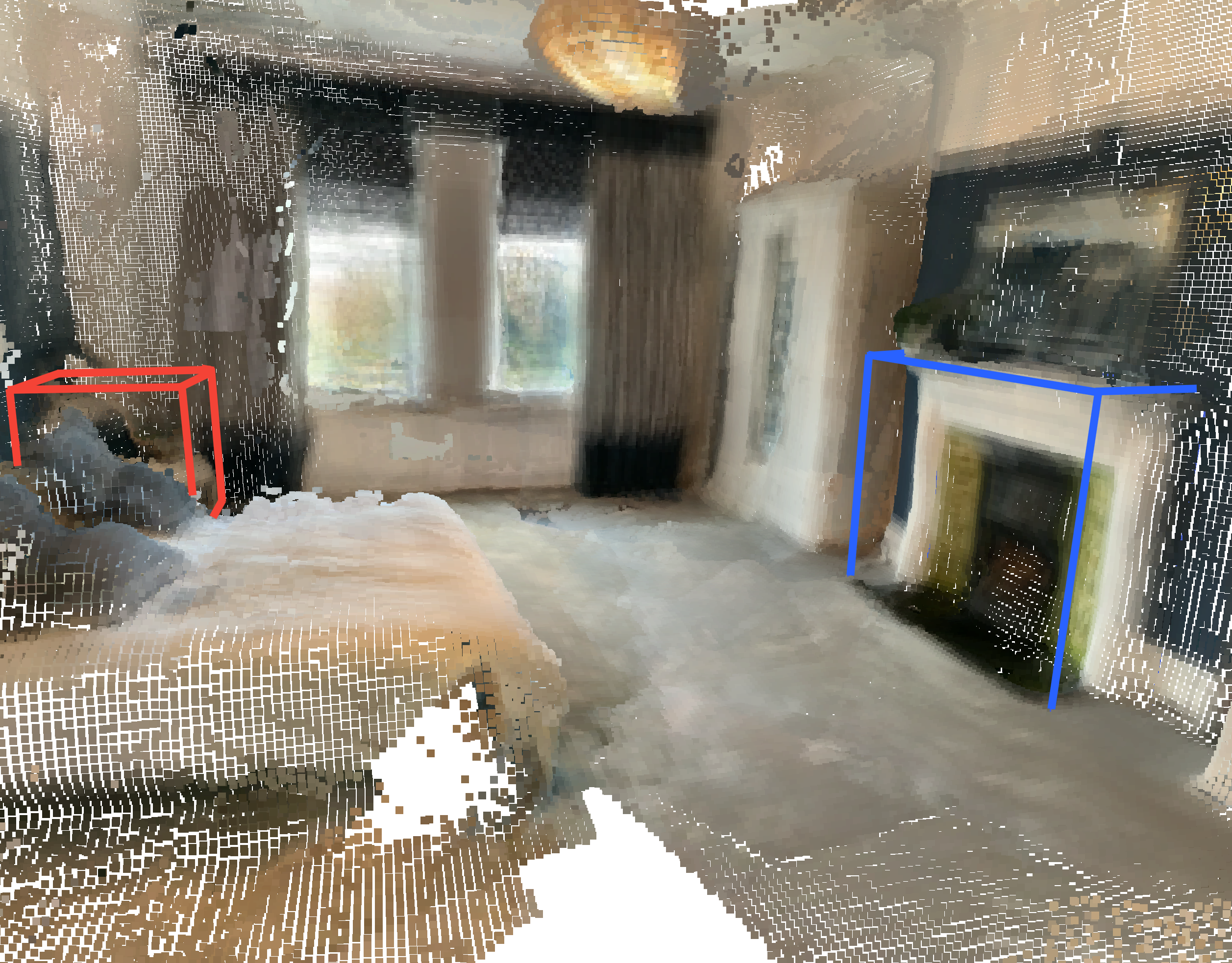} & \makecell[l]{\textit{a photo of} \\ \textcolor[rgb]{0.84,0.32,0.26}{\textit{chair}} \\ \textcolor[rgb]{0.35,0.47,0.89}{\textit{fireplace}}}
\end{tabular}
}
\caption{Qualitative results of \ours\textsubscript{0} on ScanNet++ (top row) and ARKitScenes (bottom row).}
\label{fig:qualitative-zoo3d0-scannetpp-arkit}
\end{figure}

\section{Qualitative Results} \label{sec:supp-visualization}

Open-vocabulary 3D object detection results on ScanNet200 are shown in Fig.~\ref{fig:qualitative-zoo3d0-scannet200}, while results on ARKitScenes and ScanNet++ are presented in Fig.~\ref{fig:qualitative-zoo3d0-scannetpp-arkit}. We visualize predictions from different input modalities to provide intuition of how the prediction accuracy depends on the amount of information passed to the model. 

\begin{figure}[h!]
\centering \small
\resizebox{\linewidth}{!}{
\begin{tabular}{ccc}
\multirow{2}{*}{Ground Truth} & \multicolumn{1}{c}{\ours\textsubscript{0} predictions from} & \multirow{2}{*}{Text prompt} \\
& Unposed RGB \\
\includegraphics[width=0.4\linewidth, valign=c]{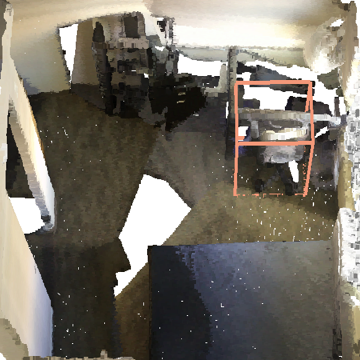} & 
\includegraphics[width=0.4\linewidth, valign=c]{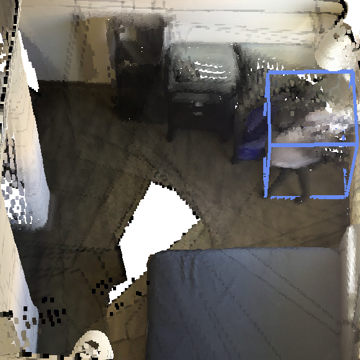} & \makecell[l]{\textit{a photo of} \\ \textcolor[rgb]{0.96,0.6,0.48}{\textit{chair}} \\ \textcolor[rgb]{0.55,0.69,1.}{\textit{office chair}}} \\
& Posed RGB & \\
\includegraphics[width=0.4\linewidth, valign=c]{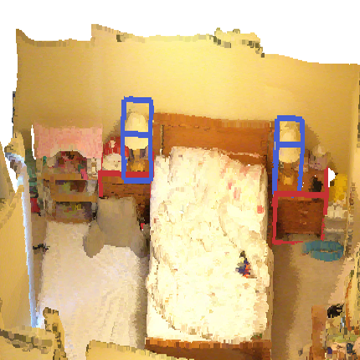} & 
\includegraphics[width=0.4\linewidth, valign=c]{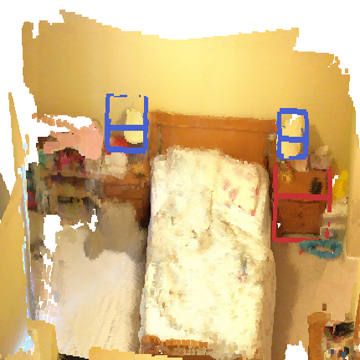} & \makecell[l]{\textit{a photo of} \\ \textcolor[rgb]{0.29,0.38,0.82}{\textit{lamp}} \\ \textcolor[rgb]{0.77,0.2,0.2}{\textit{nightstand}}} \\
\end{tabular}
}
\caption{Failure cases of \ours\textsubscript{0} on ScanNet200.}
\label{fig:failure}
\end{figure}

\paragraph{Failure cases} are depicted in Fig.~\ref{fig:failure}. Here we challenge our model in restrictive posed and unposed image-based scenarios. As metric values suggest, our model is prone to errors of various types: it can misclassify detected objects or even miss them in the scene entirely.

{
    \small
    \bibliographystyle{ieeenat_fullname}
    \bibliography{main}

@article{yang2023sam3d,
  title={Sam3d: Segment anything in 3d scenes},
  author={Yang, Yunhan and Wu, Xiaoyang and He, Tong and Zhao, Hengshuang and Liu, Xihui},
  journal={arXiv preprint arXiv:2306.03908},
  year={2023}
}

@article{zhang2023sam3d-outdoor,
  title={Sam3d: Zero-shot 3d object detection via segment anything model},
  author={Zhang, Dingyuan and Liang, Dingkang and Yang, Hongcheng and Zou, Zhikang and Ye, Xiaoqing and Liu, Zhe and Bai, Xiang},
  journal={arXiv preprint arXiv:2306.02245},
  year={2023}
}

@inproceedings{hsu2025openm3d,
  title={OpenM3D: Open Vocabulary Multi-view Indoor 3D Object Detection without Human Annotations},
  author={Hsu, Peng-Hao and Zhang, Ke and Wang, Fu-En and Tu, Tao and Li, Ming-Feng and Liu, Yu-Lun and Chen, Albert YC and Sun, Min and Kuo, Cheng-Hao},
  booktitle={Proceedings of the IEEE/CVF International Conference on Computer Vision},
  pages={8688--8698},
  year={2025}
}

@inproceedings{wang2024ov-uni3detr,
  title={Ov-uni3detr: Towards unified open-vocabulary 3d object detection via cycle-modality propagation},
  author={Wang, Zhenyu and Li, Yali and Liu, Taichi and Zhao, Hengshuang and Wang, Shengjin},
  booktitle={European Conference on Computer Vision},
  pages={73--89},
  year={2024},
  organization={Springer}
}

@article{guo2025seed1.5-vl,
  title={Seed1. 5-vl technical report},
  author={Guo, Dong and Wu, Faming and Zhu, Feida and Leng, Fuxing and Shi, Guang and Chen, Haobin and Fan, Haoqi and Wang, Jian and Jiang, Jianyu and Wang, Jiawei and others},
  journal={arXiv preprint arXiv:2505.07062},
  year={2025}
}

@article{bai2025qwen2.5-vl,
  title={Qwen2. 5-vl technical report},
  author={Bai, Shuai and Chen, Keqin and Liu, Xuejing and Wang, Jialin and Ge, Wenbin and Song, Sibo and Dang, Kai and Wang, Peng and Wang, Shijie and Tang, Jun and others},
  journal={arXiv preprint arXiv:2502.13923},
  year={2025}
}

@inproceedings{lu2023ov-3det,
  title={Open-vocabulary point-cloud object detection without 3d annotation},
  author={Lu, Yuheng and Xu, Chenfeng and Wei, Xiaobao and Xie, Xiaodong and Tomizuka, Masayoshi and Keutzer, Kurt and Zhang, Shanghang},
  booktitle={Proceedings of the IEEE/CVF conference on computer vision and pattern recognition},
  pages={1190--1199},
  year={2023}
}

@article{cao2023coda,
  title={Coda: Collaborative novel box discovery and cross-modal alignment for open-vocabulary 3d object detection},
  author={Cao, Yang and Yihan, Zeng and Xu, Hang and Xu, Dan},
  journal={Advances in Neural Information Processing Systems},
  volume={36},
  pages={71862--71873},
  year={2023}
}

@inproceedings{peng2024glis,
  title={Global-Local Collaborative Inference with LLM for Lidar-Based Open-Vocabulary Detection},
  author={Peng, Xingyu and Bai, Yan and Gao, Chen and Yang, Lirong and Xia, Fei and Mu, Beipeng and Wang, Xiaofei and Liu, Si},
  booktitle={European Conference on Computer Vision},
  pages={367--384},
  year={2024},
  organization={Springer}
}

@article{yang2024imov3d,
  title={ImOV3D: Learning Open Vocabulary Point Clouds 3D Object Detection from Only 2D Images},
  author={Yang, Timing and Ju, Yuanliang and Yi, Li},
  journal={Advances in Neural Information Processing Systems},
  volume={37},
  pages={141261--141291},
  year={2024}
}

@inproceedings{zhu2023pointclip,
  title={Pointclip v2: Prompting clip and gpt for powerful 3d open-world learning},
  author={Zhu, Xiangyang and Zhang, Renrui and He, Bowei and Guo, Ziyu and Zeng, Ziyao and Qin, Zipeng and Zhang, Shanghang and Gao, Peng},
  booktitle={Proceedings of the IEEE/CVF international conference on computer vision},
  pages={2639--2650},
  year={2023}
}

@inproceedings{jiao2024inha,
  title={Unlocking textual and visual wisdom: Open-vocabulary 3d object detection enhanced by comprehensive guidance from text and image},
  author={Jiao, Pengkun and Zhao, Na and Chen, Jingjing and Jiang, Yu-Gang},
  booktitle={European Conference on Computer Vision},
  pages={376--392},
  year={2024},
  organization={Springer}
}

@inproceedings{kolodiazhnyi2025unidet3d,
  title={Unidet3d: Multi-dataset indoor 3d object detection},
  author={Kolodiazhnyi, Maksim and Vorontsova, Anna and Skripkin, Matvey and Rukhovich, Danila and Konushin, Anton},
  booktitle={Proceedings of the AAAI Conference on Artificial Intelligence},
  volume={39},
  number={4},
  pages={4365--4373},
  year={2025}
}

@inproceedings{wang2024dust3r,
  title={Dust3r: Geometric 3d vision made easy},
  author={Wang, Shuzhe and Leroy, Vincent and Cabon, Yohann and Chidlovskii, Boris and Revaud, Jerome},
  booktitle={Proceedings of the IEEE/CVF Conference on Computer Vision and Pattern Recognition},
  pages={20697--20709},
  year={2024}
}

@article{lu2022ov-3detic,
  title={Open-vocabulary 3d detection via image-level class and debiased cross-modal contrastive learning},
  author={Lu, Yuheng and Xu, Chenfeng and Wei, Xiaobao and Xie, Xiaodong and Tomizuka, Masayoshi and Keutzer, Kurt and Zhang, Shanghang},
  journal={arXiv preprint arXiv:2207.01987},
  year={2022}
}

@inproceedings{zhang2024fm-ov3d,
  title={Fm-ov3d: Foundation model-based cross-modal knowledge blending for open-vocabulary 3d detection},
  author={Zhang, Dongmei and Li, Chang and Zhang, Renrui and Xie, Shenghao and Xue, Wei and Xie, Xiaodong and Zhang, Shanghang},
  booktitle={Proceedings of the AAAI Conference on Artificial Intelligence},
  volume={38},
  number={15},
  pages={16723--16731},
  year={2024}
}

@article{zhu2023object2scene,
  title={Object2scene: Putting objects in context for open-vocabulary 3d detection},
  author={Zhu, Chenming and Zhang, Wenwei and Wang, Tai and Liu, Xihui and Chen, Kai},
  journal={arXiv preprint arXiv:2309.09456},
  year={2023}
}

@article{teed2021droid,
  title={Droid-slam: Deep visual slam for monocular, stereo, and rgb-d cameras},
  author={Teed, Zachary and Deng, Jia},
  journal={Advances in neural information processing systems},
  volume={34},
  pages={16558--16569},
  year={2021}
}

@inproceedings{wang2025vggt,
  title={Vggt: Visual geometry grounded transformer},
  author={Wang, Jianyuan and Chen, Minghao and Karaev, Nikita and Vedaldi, Andrea and Rupprecht, Christian and Novotny, David},
  booktitle={Proceedings of the Computer Vision and Pattern Recognition Conference},
  pages={5294--5306},
  year={2025}
}

@inproceedings{radford2021learning,
  title={Learning transferable visual models from natural language supervision},
  author={Radford, Alec and Kim, Jong Wook and Hallacy, Chris and Ramesh, Aditya and Goh, Gabriel and Agarwal, Sandhini and Sastry, Girish and Askell, Amanda and Mishkin, Pamela and Clark, Jack and others},
  booktitle={International conference on machine learning},
  pages={8748--8763},
  year={2021},
  organization={PmLR}
}

@inproceedings{rukhovich2023tr3d,
  title={Tr3d: Towards real-time indoor 3d object detection},
  author={Rukhovich, Danila and Vorontsova, Anna and Konushin, Anton},
  booktitle={2023 IEEE International Conference on Image Processing (ICIP)},
  pages={281--285},
  year={2023},
  organization={IEEE}
}

@inproceedings{rukhovich2022fcaf3d,
  title={Fcaf3d: Fully convolutional anchor-free 3d object detection},
  author={Rukhovich, Danila and Vorontsova, Anna and Konushin, Anton},
  booktitle={European Conference on Computer Vision},
  pages={477--493},
  year={2022},
  organization={Springer}
}

@inproceedings{yan2024maskclustering,
  title={Maskclustering: View consensus based mask graph clustering for open-vocabulary 3d instance segmentation},
  author={Yan, Mi and Zhang, Jiazhao and Zhu, Yan and Wang, He},
  booktitle={Proceedings of the IEEE/CVF Conference on Computer Vision and Pattern Recognition},
  pages={28274--28284},
  year={2024}
}

@article{ravi2024sam2,
  title={Sam 2: Segment anything in images and videos},
  author={Ravi, Nikhila and Gabeur, Valentin and Hu, Yuan-Ting and Hu, Ronghang and Ryali, Chaitanya and Ma, Tengyu and Khedr, Haitham and R{\"a}dle, Roman and Rolland, Chloe and Gustafson, Laura and others},
  journal={arXiv preprint arXiv:2408.00714},
  year={2024}
}

@article{shen2023v-detr,
  title={V-detr: Detr with vertex relative position encoding for 3d object detection},
  author={Shen, Yichao and Geng, Zigang and Yuan, Yuhui and Lin, Yutong and Liu, Ze and Wang, Chunyu and Hu, Han and Zheng, Nanning and Guo, Baining},
  journal={arXiv preprint arXiv:2308.04409},
  year={2023}
}

@inproceedings{zheng2025video-3d-llm,
  title={Video-3d llm: Learning position-aware video representation for 3d scene understanding},
  author={Zheng, Duo and Huang, Shijia and Wang, Liwei},
  booktitle={Proceedings of the Computer Vision and Pattern Recognition Conference},
  pages={8995--9006},
  year={2025}
}

@inproceedings{zhi2025lscenellm,
  title={Lscenellm: Enhancing large 3d scene understanding using adaptive visual preferences},
  author={Zhi, Hongyan and Chen, Peihao and Li, Junyan and Ma, Shuailei and Sun, Xinyu and Xiang, Tianhang and Lei, Yinjie and Tan, Mingkui and Gan, Chuang},
  booktitle={Proceedings of the Computer Vision and Pattern Recognition Conference},
  pages={3761--3771},
  year={2025}
}

@article{huang2024chat-scene,
  title={Chat-scene: Bridging 3d scene and large language models with object identifiers},
  author={Huang, Haifeng and Chen, Yilun and Wang, Zehan and Huang, Rongjie and Xu, Runsen and Wang, Tai and Liu, Luping and Cheng, Xize and Zhao, Yang and Pang, Jiangmiao and others},
  journal={Advances in Neural Information Processing Systems},
  volume={37},
  pages={113991--114017},
  year={2024}
}

@inproceedings{tang2025onlineanyseg,
  title={Onlineanyseg: Online zero-shot 3d segmentation by visual foundation model guided 2d mask merging},
  author={Tang, Yijie and Zhang, Jiazhao and Lan, Yuqing and Guo, Yulan and Dong, Dezun and Zhu, Chenyang and Xu, Kai},
  booktitle={Proceedings of the Computer Vision and Pattern Recognition Conference},
  pages={3676--3685},
  year={2025}
}

@inproceedings{zhao2025sam2object,
  title={SAM2Object: Consolidating View Consistency via SAM2 for Zero-Shot 3D Instance Segmentation},
  author={Zhao, Jihuai and Zhuo, Junbao and Chen, Jiansheng and Ma, Huimin},
  booktitle={Proceedings of the Computer Vision and Pattern Recognition Conference},
  pages={19325--19334},
  year={2025}
}

@inproceedings{rukhovich2022imvoxelnet,
  title={Imvoxelnet: Image to voxels projection for monocular and multi-view general-purpose 3d object detection},
  author={Rukhovich, Danila and Vorontsova, Anna and Konushin, Anton},
  booktitle={Proceedings of the IEEE/CVF winter conference on applications of computer vision},
  pages={2397--2406},
  year={2022}
}

@inproceedings{tu2023imgeonet,
  title={Imgeonet: Image-induced geometry-aware voxel representation for multi-view 3d object detection},
  author={Tu, Tao and Chuang, Shun-Po and Liu, Yu-Lun and Sun, Cheng and Zhang, Ke and Roy, Donna and Kuo, Cheng-Hao and Sun, Min},
  booktitle={Proceedings of the IEEE/CVF International Conference on Computer Vision},
  pages={6996--7007},
  year={2023}
}

@article{huang2025nerf-det++,
  title={Nerf-det++: Incorporating semantic cues and perspective-aware depth supervision for indoor multi-view 3d detection},
  author={Huang, Chenxi and Hou, Yuenan and Ye, Weicai and Huang, Di and Huang, Xiaoshui and Lin, Binbin and Cai, Deng},
  journal={IEEE Transactions on Image Processing},
  year={2025},
  publisher={IEEE}
}

@article{zhu2024llava-3d,
  title={Llava-3d: A simple yet effective pathway to empowering lmms with 3d-awareness},
  author={Zhu, Chenming and Wang, Tai and Zhang, Wenwei and Pang, Jiangmiao and Liu, Xihui},
  journal={arXiv preprint arXiv:2409.18125},
  year={2024}
}

@article{qi2025gpt4scene,
  title={Gpt4scene: Understand 3d scenes from videos with vision-language models},
  author={Qi, Zhangyang and Zhang, Zhixiong and Fang, Ye and Wang, Jiaqi and Zhao, Hengshuang},
  journal={arXiv preprint arXiv:2501.01428},
  year={2025}
}

@article{fan2025vlm-3r,
  title={VLM-3R: Vision-Language Models Augmented with Instruction-Aligned 3D Reconstruction},
  author={Fan, Zhiwen and Zhang, Jian and Li, Renjie and Zhang, Junge and Chen, Runjin and Hu, Hezhen and Wang, Kevin and Qu, Huaizhi and Wang, Dilin and Yan, Zhicheng and others},
  journal={arXiv preprint arXiv:2505.20279},
  year={2025}
}

@article{mao2025spatiallm,
  title={SpatialLM: Training Large Language Models for Structured Indoor Modeling},
  author={Mao, Yongsen and Zhong, Junhao and Fang, Chuan and Zheng, Jia and Tang, Rui and Zhu, Hao and Tan, Ping and Zhou, Zihan},
  journal={arXiv preprint arXiv:2506.07491},
  year={2025}
}

@inproceedings{dai2017scannet,
  title={Scannet: Richly-annotated 3d reconstructions of indoor scenes},
  author={Dai, Angela and Chang, Angel X and Savva, Manolis and Halber, Maciej and Funkhouser, Thomas and Nie{\ss}ner, Matthias},
  booktitle={Proceedings of the IEEE conference on computer vision and pattern recognition},
  pages={5828--5839},
  year={2017}
}

@article{baruch2021arkitscenes,
  title={Arkitscenes: A diverse real-world dataset for 3d indoor scene understanding using mobile rgb-d data},
  author={Baruch, Gilad and Chen, Zhuoyuan and Dehghan, Afshin and Dimry, Tal and Feigin, Yuri and Fu, Peter and Gebauer, Thomas and Joffe, Brandon and Kurz, Daniel and Schwartz, Arik and others},
  journal={arXiv preprint arXiv:2111.08897},
  year={2021}
}

@inproceedings{rozenberszki2022scannet200,
  title={Language-grounded indoor 3d semantic segmentation in the wild},
  author={Rozenberszki, David and Litany, Or and Dai, Angela},
  booktitle={European conference on computer vision},
  pages={125--141},
  year={2022},
  organization={Springer}
}

@inproceedings{yeshwanth2023scannet++,
  title={Scannet++: A high-fidelity dataset of 3d indoor scenes},
  author={Yeshwanth, Chandan and Liu, Yueh-Cheng and Nie{\ss}ner, Matthias and Dai, Angela},
  booktitle={Proceedings of the IEEE/CVF International Conference on Computer Vision},
  pages={12--22},
  year={2023}
}

@inproceedings{qi2019votenet,
  title={Deep hough voting for 3d object detection in point clouds},
  author={Qi, Charles R and Litany, Or and He, Kaiming and Guibas, Leonidas J},
  booktitle={proceedings of the IEEE/CVF International Conference on Computer Vision},
  pages={9277--9286},
  year={2019}
}

@inproceedings{liu2024grounding-dino,
  title={Grounding dino: Marrying dino with grounded pre-training for open-set object detection},
  author={Liu, Shilong and Zeng, Zhaoyang and Ren, Tianhe and Li, Feng and Zhang, Hao and Yang, Jie and Jiang, Qing and Li, Chunyuan and Yang, Jianwei and Su, Hang and others},
  booktitle={European conference on computer vision},
  pages={38--55},
  year={2024},
  organization={Springer}
}

@article{ren2024grounded-sam,
  title={Grounded sam: Assembling open-world models for diverse visual tasks},
  author={Ren, Tianhe and Liu, Shilong and Zeng, Ailing and Lin, Jing and Li, Kunchang and Cao, He and Chen, Jiayu and Huang, Xinyu and Chen, Yukang and Yan, Feng and others},
  journal={arXiv preprint arXiv:2401.14159},
  year={2024}
}

@article{yao2024ovmono3d,
  title={Open vocabulary monocular 3d object detection},
  author={Yao, Jin and Gu, Hao and Chen, Xuweiyi and Wang, Jiayun and Cheng, Zezhou},
  journal={arXiv preprint arXiv:2411.16833},
  year={2024}
}

@article{yao2025labelany3d,
  title={LabelAny3D: Label Any Object 3D in the Wild},
  author={Yao, Jin and Redoy, Radowan Mahmud and Elbaum, Sebastian and Dwyer, Matthew and Cheng, Zezhou},
  journal={Advances in neural information processing systems},
  year={2025}
}
}

\end{document}